\documentclass[10pt,twocolumn,letterpaper]{article}

\usepackage{iccv}              
\usepackage{multirow}
\usepackage{diagbox} 
\usepackage[accsupp]{axessibility}

\newcommand{\myparagraph}[1]{\smallskip\noindent\textbf{#1}}

\definecolor{iccvblue}{rgb}{0.21,0.49,0.74}
\usepackage[pagebackref,breaklinks,colorlinks,allcolors=iccvblue]{hyperref}

\def\paperID{10461} 
\def\confName{ICCV}
\def\confYear{2025}

\title{\emph{Learn2Synth}: Learning Optimal Data Synthesis Using Hypergradients\\for Brain Image Segmentation}

\author{
  Xiaoling Hu$^{1,\dagger}$, Xiangrui Zeng$^{1}$, Oula Puonti$^{1,2}$, \\
  Juan Eugenio Iglesias$^{1,3,4}$, Bruce Fischl$^{1,\ddagger}$, Yaël Balbastre$^{1,5,\ddagger}$ \\
  \\
  $^{1}$Massachusetts General Hospital and Harvard Medical School \\ 
  $^{2}$Danish Research Centre for Magnetic Resonance, Copenhagen University Hospital \\
  $^{3}$Centre for Medical Image Computing, University College London \\
  $^{4}$Computer Science and AI Laboratory, Massachusetts Institute of Technology \\
  $^{5}$Department of Experimental Psychology, University College London \\
}

\begin{document}

\maketitle
\renewcommand{\thefootnote}{\fnsymbol{footnote}}
\footnotetext{$\dagger$ Email: Xiaoling Hu (xihu3@mgh.harvard.edu);~$\ddagger$ Co-senior authors}
\renewcommand{\thefootnote}{\arabic{footnote}}
\begin{figure*}[ht!]
\centering
    \includegraphics[width=1\textwidth]{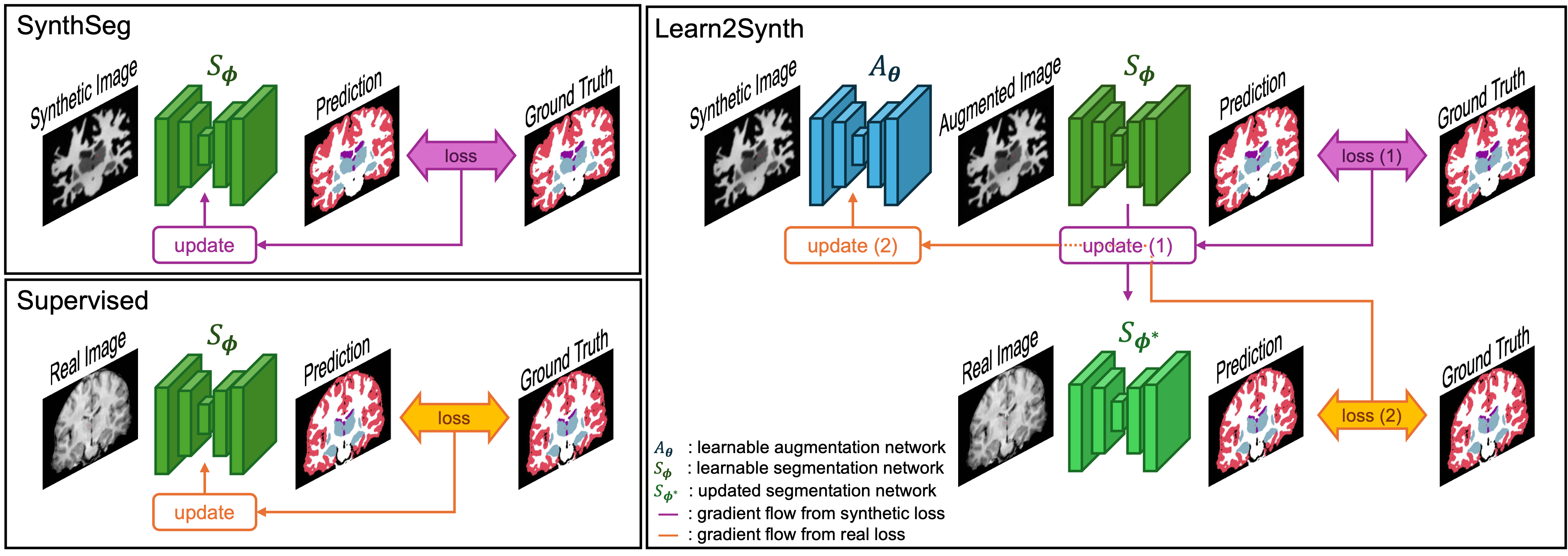}
    \caption{\emph{SynthSeg}~\cite{billot2023synthseg} (top left) is a domain randomization strategy that has shown great success in neuroimaging applications, where an image with random contrast is synthesized from a label map using a set of stochastic rules. This strategy contrasts with typical supervised training~\cite{milletari2016v} (bottom left), where a model is trained on real labeled images (possibly augmented). Our proposed strategy, \emph{Learn2Synth} (right), interleaves a typical \emph{SynthSeg} step \textcolor{purple}{\textbf{(1)}}, that optimizes a segmentation network (\textcolor{OliveGreen}{$S_{\boldsymbol{\phi}}$}) on synthetic data, with a supervised step \textcolor{orange}{\textbf{(2)}}. Importantly, the supervised step is not used to update the segmentation network, but to update an augmentation function (\textcolor{NavyBlue}{$A_{\boldsymbol{\theta}}$}). This is made possible by backpropagating the supervised loss \textcolor{orange}{\textbf{(2)}} through the update step \textcolor{purple}{\textbf{(1)}}.
    }
    \label{fig:intro}
\end{figure*}

\begin{abstract}
Domain randomization through synthesis is a powerful strategy to train networks that are unbiased with respect to the domain of the input images. Randomization allows networks to see a virtually infinite range of intensities and artifacts during training, thereby minimizing overfitting to appearance and maximizing generalization to unseen data. Although powerful, this approach relies on the accurate tuning of a large set of hyperparameters that govern the probabilistic distribution of the synthesized images. Instead of manually tuning these parameters, we introduce \emph{Learn2Synth}, a novel procedure in which synthesis parameters are learned using a small set of real labeled data. Unlike methods that impose constraints to align synthetic data with real data (\emph{e.g.}, contrastive or adversarial techniques), which risk misaligning the image and its label map, we tune an augmentation engine such that a segmentation network trained on synthetic data has optimal accuracy when applied to real data. This approach allows the training procedure to benefit from real labeled examples, without ever using these real examples to train the segmentation network, which avoids biasing the network towards the properties of the training set. Specifically, we develop parametric and nonparametric strategies to enhance synthetic images in a way that improves the performance of the segmentation network. We demonstrate the effectiveness of this learning strategy on synthetic and real-world brain scans. Code is available at: \url{https://github.com/HuXiaoling/Learn2Synth}.
\end{abstract}
\section{Introduction}


In medical imaging, the availability of high-quality reference data (high-resolution and artifact-free images, expert annotations, fair representation of populations) is a major limiting factor, compared to other computer vision domains~\cite{varoquaux2022machine}. This can be explained by the cost of data acquisition, noise and artifacts inherent to medical images, as well as the time and expertise required to label dense three-dimensional images. This limitation has hampered the development of modality-agnostic tools and led to a web of highly specific models trained on small datasets, whose performance degrades greatly in the presence of slight domain shifts~\cite{jog2018pulse,varoquaux2018cross, maleki2022generalizability}, making out-of-site generalizability one of the most difficult and pervasive problems in biomedical machine learning~\cite{bluemke2020assessing}. This behavior, specific to supervised training on small datasets -- peak on-domain performance with a rapid decay as data shifts away from the training domain -- has long hindered the adoption of learning-based methods by the academic and clinical community. In contrast, supervised shallow models and unsupervised Bayesian approaches reach moderate but consistent performances across the board~\cite{schulz2020different,baldwin2022deep}. 

In a data-limited regime, randomized augmentation remains crucial and provides performance boosts in most contexts~\cite{taylor2018improving,shorten2019survey,yarats2021image}. The use of proper augmentation -- along with methodical architecture and hyperparameter tuning -- has been shown to be one of the main predictors of accuracy on an array of medical image segmentation challenges~\cite{antonelli2022medical,isensee2021nnu}. The introduction of domain randomization~\cite{tobin2017domain, bengio2011deep, zhao2019data, tremblay2018training}, in which the concept of augmentation is pushed even further by randomly sampling entire images, has been a paradigm shift in this regard. In its version most commonly used in biomedical imaging, domain randomization starts with a set of densely annotated shapes from which intensities are conditionally generated, followed by a set of common-but-extreme image augmentation steps~\cite{billot2023synthseg}. This strategy ensures that labels and images are perfectly aligned by construction. Nonetheless, performance and generalizability are often sensitive to the choice of hyper-parameters that govern the statistical distributions of augmentations, which must be carefully validated on a dataset of real data. Furthermore, the synthesis procedure relies on \emph{ad hoc} rules that may not capture the full range of features observed in real images, thereby setting a ceiling in achievable accuracy. Indeed, models trained with synthetic data always perform slightly worse than supervised models applied on-domain (but generalize significantly better)~\cite{billot2023synthseg} due to the so-called ``reality gap''~\cite{jakobi1995noise}. 

Optimally, one would like to keep the best of both approaches, namely high on-domain accuracy and high generalizability to off-domain data. A simple solution consists of training models on a mix of synthetic and real data, sampled with some predefined proportion~\cite{nikolenko2021synthetic, muller2019does}. However, it is possible to train a single network in such a way that it ends up containing multiple non-overlapping subnetworks~\cite{zhang2021ex}, and it is likely that naively-trained deep models also internalize parallel subnetworks. Models trained on both synthetic and real data therefore risk overfitting some of their filters to the small set of real data. An alternative approach consists of generating realistic data from labels using losses that aim to match the distributions of synthetic and real images (adversarial~\cite{shrivastava2017learning, ganin2018synthesizing}, contrastive~\cite{chen2020simple, he2020momentum}, normalizing flows~\cite{kingma2018glow, papamakarios2017masked}, diffusion~\cite{ho2020denoising, nichol2021improved}). This however introduces an objective that is not relevant to the original task, leading to two major drawbacks. First, conditional labels and generated features are not constrained to be aligned, especially if the labels’ distribution deviates from the distribution of their target features in the real images~\cite{liu2022undoing}. Second, synthetic pairs may perceptually match real images without being able to train an accurate segmentation network. Indeed, distribution-matching models may spend too much capacity synthesizing features that have no impact on segmentation accuracy, or conversely fail to generate off-domain features that induce segmentation robustness.

Instead, we introduce \emph{Learn2Synth}, a model trained with a single objective: improving the \emph{real world accuracy} of a network \emph{trained on synthetic data}  (\Cref{fig:intro}). To do so, we proceed in two steps: \textbf{(1)} we freeze the synthesis network, push synthetic data through the segmentation network and update it; \textbf{(2)} we freeze the segmentation network, push real data through the segmentation network, and update the synthesis network. Naively, if a network is fed real data, no gradients flow back to the synthesis network; the key to our approach is to leverage hyper-gradients~\cite{maclaurin2015gradient,baydin2017online,franceschi2017forward,chandra2022gradient}, obtained by differentiating through step \textbf{(1)}. In summary:
\begin{enumerate}
    \item We propose a novel training pipeline for segmentation tasks where synthetic data, generated under \emph{ad hoc} rules, is augmented by a trainable network. This avoids direct reliance on real data for training the segmentation network while still improving segmentation accuracy.

    \item Instead of promoting realism using adversarial or perceptual loss functions, our strategy encourages the augmentation network to generate synthetic samples that optimize segmentation performance \textit{on real data}.

    \item By alternating between synthetic and real passes, the method improves the robustness and generalizability of the segmentation network, enhancing its performance across different image contrasts while preserving data-label pair integrity.

    \item We demonstrate that our strategy improves performance across the board compared to pure domain randomization, while greatly improving generalizability compared to supervised training. Furthermore, the learned augmentation parameters give us insight into the optimal environment for network training.
\end{enumerate}

\section{Related work}

\myparagraph{Learning-based augmentation.} Deep learning methods are prone to overfitting, especially when training data are limited. Data augmentation is a widely used technique to counteract this issue by artificially expanding the diversity of the training dataset. Traditional data augmentation techniques involve transformations such as intensity adjustments, spatial distortions, and geometric alterations~\cite{zhang2020generalizing}. These transformations help neural networks generalize better by exposing them to a wider range of variations. However, they rely on handcrafted transformations and hyperparameters that may not be optimal for specific datasets or tasks. Learning-based augmentation methods aim to alleviate this problem; they leverage deep learning models to learn optimal data transformations based on the dataset at hand, rather than relying on predefined parameters. One of the pioneering works in this area is AutoAugment~\cite{cubuk2019autoaugment}, which uses reinforcement learning to identify the best augmentation policies for a given dataset, while Fast AutoAugment~\cite{lim2019fast} and Faster AutoAugment~\cite{hataya2020faster} improve efficiency and scalability in the search process. In biomedical imaging, one typically has access to a large set of unlabeled data but a very small (or partial) set of labels. In this context, adversarial augmentation has been used to optimize augmentation parameters dynamically during training~\cite{zhang2017deep, chaitanya2019semi, chen2022enhancing}, producing challenging examples that encourage the model to learn more robust features. In fully unsupervised settings, contrastive learning has shown promise by generating multiple augmented views of the same image to capture invariant representations~\cite{chaitanya2020contrastive, you2022simcvd}, thereby leveraging unlabeled data to improve feature extraction and generalization. Alternatively, CycleGAN~\cite{zhu2017unpaired} and other simpler GAN-based methods have been used to translate labels to images and images to labels without requiring explicit pairing \cite{sindel2023vesselsegmentation, al2020learning}.

\myparagraph{Domain adaptation.} A model trained on one domain usually does not perform well on images from another domain. Domain adaptation is a way to address the challenge and explicitly seeks to bridge the gaps between a source domain with labeled data, and a target domain without labels. One of the common solutions is to map both source and target domains to a common latent space~\cite{kamnitsas2017unsupervised, dou2019pnp, ganin2016domain, you2022incremental, wang2018deep}. By using image-to-image translation methods, generative adaptation methods have tried to match the source domains to the target domain~\cite{sandfort2019data, huo2018synseg, zhang2018translating}. Researchers have also proposed to operate in both feature and image space, leading to impressive results in cross-modality segmentation~\cite{chen2019synergistic, hoffman2018cycada}. For intra-modality scenarios, impressive performances have been obtained with test-time adaptation methods~\cite{karani2021test, he2021autoencoder}, requiring only light fine-tuning at test-time. 


\myparagraph{Synthetic training data and domain randomization.} Large amounts of high-quality labeled data are essential for the success of deep learning methods.  It is always labor-intensive and costly to obtain high-quality annotations, especially for biomedical domains, which usually require domain knowledge and large amounts of work. Instead of using real data to train the networks, synthetic data can be introduced to increase generalization and robustness, either generated with physics-based models~\cite{jog2018pulse}, adversarial generative networks~\cite{frid2018synthetic, chartsias2017multimodal, zhang2021datasetgan, li2021semantic, sushko2023one, tritrong2021repurposing}, or conditioned on label maps~\cite{mahmood2019deep, isola2017image, billot2023synthseg}. These strategies can generate large training datasets with perfect ground truth without any cost~\cite{richter2016playing}. However, the generated synthetic images suffer from a ``reality gap''~\cite{jakobi1995noise}. In neuroimaging, domain-randomization underpins a family of contrast-invariant methods that target an array of tasks (SynthSeg~\cite{billot2023synthseg, billot2023robust}, SynthSR~\cite{iglesias2023synthsr}, SynthMorph~\cite{hoffmann2021synthmorph}, SynthStrip~\cite{hoopes2022synthstrip}). By randomizing the training data, networks can experience a virtually infinite variety of intensities and artifacts, which helps to reduce overfitting and enhance generalization to unseen data. However, this augmentation technique relies on \emph{ad hoc} rules, which can constrain the diversity of shapes and artifacts generated.


\section{Method}
\label{sec:method}

\subsection{Training procedure}
\label{sec:training}

We have designed a training procedure in which synthetic training examples generated using \emph{ad hoc} rules are enhanced by a trainable network, before being used to train a segmentation network. The segmentation network is trained \emph{only} on synthetic examples, whereas the enhancement network is trained by backpropagating a loss computed \emph{only} on real images (\Cref{fig:intro}). This strategy encourages the augmentation network to produce samples that improve the accuracy of the segmentation network on real data; leveraging real labeled examples without directly using them in training the segmentation network. In the following sections, $(\mathbf{x}_\mathrm{real}, \mathbf{y}_\mathrm{real})$ and $(\mathbf{x}_\mathrm{synth}, \mathbf{y}_\mathrm{synth})$ represent real and synthetic labeled examples, respectively, while $A_{\boldsymbol{\theta}}$ and $S_{\boldsymbol{\phi}}$ denote the augmentation and segmentation functions with corresponding weights $\boldsymbol{\theta}$ and $\boldsymbol{\phi}$. Our method consists of a \emph{synthetic pass} and a \emph{real pass}, which we describe in detail.

\subsubsection{Synthetic pass}

In the \emph{synthetic pass}, a synthetic image, generated from a label map, is first pushed through an augmentation function $A_{\boldsymbol{\theta}}$. Its output is denoted as $A_{\boldsymbol\theta}(\mathbf{x}_\mathrm{synth})$, and the pair $(A_{\boldsymbol\theta}(\mathbf{x}_\mathrm{synth}), \mathbf{y}_\mathrm{synth})$ is subsequently used to train the segmentation network. The forward pass applied to the synthetic data yields
\begin{equation}
    \mathcal{L}_\mathrm{synth} \triangleq \operatorname{SoftDice}\left(S_{\boldsymbol\phi}\left( A_{\boldsymbol\theta}\left(\mathbf{x}_\mathrm{synth}\right)\right), \mathbf{y}_\mathrm{synth}\right),
\end{equation}
where $\operatorname{SoftDice}$ is the soft Dice loss function~\cite{sudre2017generalised} and $\mathcal{L}_\mathrm{synth}$ represents the loss value. The gradient of this loss with respect to the segmentation network’s weights is computed by automatic differentiation and used to update the segmentation network through the optimizer's update rule
\begin{align}
    \mathbf{g}_{\phi} &\triangleq \frac{\partial \mathcal{L}_\mathrm{synth}}{\partial \boldsymbol\phi}, & 
    \boldsymbol\phi^\ast &\triangleq \operatorname{update}\left(\boldsymbol\phi, \mathbf{g}_{\phi}\right).
\end{align}
It is important to note that during the \emph{synthetic pass} we only update the weights of the segmentation network ($S_{\boldsymbol\phi}$), while the parameters of the augmentation network ($A_{\boldsymbol\theta}$) remain fixed.

\subsubsection{Real pass}

Using only the \emph{synthetic pass}, it is unsurprising that the trained segmentation network (with parameters $S_{\boldsymbol\phi}$) performs well on images generated by $A_{\boldsymbol\theta}(\mathbf{x}_\mathrm{synth})$. The next question is: how can we bridge the gap between $A_{\boldsymbol\theta}(\mathbf{x}_\mathrm{synth})$ and $\mathbf{x}_\mathrm{real}$ to ensure the segmentation network also performs well on $\mathbf{x}_\mathrm{real}$? Our solution is to leverage real-labeled data by introducing a \emph{real pass} that encourages a change in $A_{\boldsymbol\theta}$ that increases the segmentation accuracy on $\mathbf{x}_\mathrm{real}$.

A real image $\mathbf{x}_\mathrm{real}$ is pushed through the (updated) segmentation network, resulting in the loss
\begin{equation}
    \mathcal{L}_\mathrm{real} \triangleq \operatorname{SoftDice}\left(S_{\boldsymbol\phi^\ast}\left( \mathbf{x}_\mathrm{real}\right), \mathbf{y}_\mathrm{real}\right).
\end{equation}
If implemented naively, no gradients would flow back into the augmentation network during the \emph{real pass} because the real image does not pass through the augmentation network. To make this strategy effective, we implement a differentiable update step that encourages the augmentation network to generate synthetic images that improve the update of the segmentation network.  The gradient of this loss with respect to the augmentation network is:
 \begin{align}
    \mathbf{g}_{\theta} \triangleq \frac{\partial \mathcal{L}_\mathrm{real}}{\partial \boldsymbol\theta} & {}= \frac{\partial \mathcal{L}_\mathrm{real}}{\partial \boldsymbol\phi^\ast} \times \frac{\partial \boldsymbol\phi^\ast}{\partial \mathbf{g}_{\phi}} \times \frac{\partial \mathbf{g}_{\phi}}{\partial \boldsymbol\theta} \nonumber\\ & {}=
    \frac{\partial \mathcal{L}_\mathrm{real}}{\partial \boldsymbol\phi^\ast} \times \frac{\partial \boldsymbol\phi^\ast}{\partial \mathbf{g}_{\phi}} \times \frac{\partial^2 \mathcal{L}_\mathrm{synth}}{\partial \boldsymbol\phi \partial \boldsymbol\theta^{T}},
 \end{align}
which is then used to update the augmentation network through the update step
\begin{equation}
    \boldsymbol\theta^\ast \triangleq \operatorname{update}(\boldsymbol\theta, \mathbf{g}_{\theta})~.
\end{equation}
There are three components in the computation of $\mathbf{g}_\theta$:
\begin{enumerate}
    \item $\frac{\partial \mathcal{L}_\mathrm{real}}{\partial \boldsymbol\phi^\ast}$ is the gradient of the real loss with respect to the current weights of the segmentation network. In a traditional training scenario using real data, this would be used to update the segmentation network's weights.
    \item $\frac{\partial \boldsymbol\phi^\ast}{\partial \mathbf{g}_\phi}$ is the gradient of the update step with respect to the previous gradient. If the update step is a gradient descent step, that term is simply an identity matrix scaled by the learning rate.
    \item $\frac{\partial^2 \mathcal{L}_\mathrm{synth}}{\partial \boldsymbol\phi \partial \boldsymbol\theta^{T}}$ is the Hessian of the synthetic loss with respect to both the segmentation and augmentation networks. This term indicates how the synthetic loss changes when taking steps in both $\boldsymbol\phi$ and $\boldsymbol\theta$.
\end{enumerate}
Note that the Hessian is only introduced for interpretation purposes. In practice, the complete gradient is obtained by automatic differentiation; the Hessian matrix is therefore never allocated in memory.



\subsection{Augmentation models}

The ``reality gap'' that leads to a decrease in performance when applying a synthetically trained model to real data can have three different causes: \textbf{(1)} a miscalibration of the synthesis hyper-parameters, \textbf{(2)} overly simplistic synthesis rules that fail to generate realistic features, or \textbf{(3)} overly complicated synthesis results that force the network to waste capacity on features that never occur in real data. We introduce two learnable augmentation functions that target these possibilities:
\begin{itemize}
    \item a parametric procedure in which the hyper-parameters that govern Gaussian noise and intensity non-uniformity are learnable (\Cref{sec:parametric});
    \item a nonparametric procedure in which a learnable enhancement network is applied to synthetic data (\Cref{sec:nonparametric}).
\end{itemize}



\subsection{Parametric model}
\label{sec:parametric}

Our parametric model encompasses two artifacts commonly found in magnetic resonance images: Gaussian noise and intensity non-uniformity (INU). For simplicity, we describe the procedure with one-dimensional signals. Its extension to two or three spatial dimensions is straightforward.



\myparagraph{INU.} Magnetic resonance images typically suffer from smooth changes in magnitude caused by the spatial profile of transmit and receive coils. While the transmit effect is nonlinear and sequence-dependent, the receive effect is multiplicative and commonly modeled using a small number of smooth basis functions. In order to make the spatial frequency of the receive field differentiable, we sample different random fields with different node spacings, which we modulate with learnable coefficients and combine. Formally, the generation of one of the random fields is written as
\begin{align}
    \boldsymbol{\beta}_k & \sim U_{M_k}(0.5, 2), & \boldsymbol{\alpha}_k & = \mathbf{B}_k\boldsymbol{\beta}_k
\end{align}
where $M_k$ is the number of low-dimensional coefficients, $\mathbf{B}_k \in \mathbb{R}^{N \times M_k}$ is a first-order B-spline basis functions with node spacing $(N-1)/(M_k-1)$ and $N$ is the number of voxels in the image. The receive field and intensity-modulated images are then constructed by
\begin{align}
    \boldsymbol{\alpha} & = \prod_{k=1}^K \boldsymbol{\alpha}_k^{c_k} ~, & 
    \mathbf{x}_\mathrm{synth} & \leftarrow \mathbf{x}_\mathrm{synth} \odot \boldsymbol{\alpha} ~,
    \label{eq:bias}
\end{align}
where $\mathbf{c}\in\mathbb{R}^{K}$ are learnable coefficients, and products are taken element-wise.


\myparagraph{Gaussian noise.} Assuming that Gaussian noise variance is identical across all images in the dataset, we implement learnable noise augmentation by sampling Gaussian noise identically distributed across voxels and scaling it by a learnable standard deviation $\sigma$, resulting in the equation
\begin{align}
\boldsymbol{\varepsilon} & \sim \mathcal{N}_{N}(0, 1), &
\mathbf{x}_\mathrm{synth} &\leftarrow \mathbf{x}_\mathrm{synth} + \sigma \cdot \boldsymbol{\varepsilon}.
\label{eq:noise}
\end{align}
However, the assumption of identical noise variance across images is very unlikely to hold in reality. Alternatively, we can learn a noise hyper-variance that is randomly modulated per-image:
\begin{align}
s \sim \mathcal{N}(0, 1), ~~~
\boldsymbol{\varepsilon} \sim \mathcal{N}_{N}(0, 1), \nonumber\\
\mathbf{x}_\mathrm{synth} \leftarrow \mathbf{x}_\mathrm{synth} + \sigma \cdot s \cdot\boldsymbol{\varepsilon}.
\label{eq:hypernoise}
\end{align}
This step ensures that the model not only learns to generate realistic noise but also maintains the ability to handle various noise levels during the augmentation process.

\myparagraph{Joint model.} The set of learnable augmentation parameters is $\boldsymbol{\theta}~=~[\mathbf{c}, \sigma]$. As described in~\Cref{sec:training}, we use the $({\mathbf{x}_\mathrm{synth}, \mathbf{y}_\mathrm{synth}})$ pairs to train the segmentation network $S_{\boldsymbol\phi}$, while the augmentation network $A_{\boldsymbol\theta}$ is responsible for generating diverse synthetic images. 

\subsection{Nonparametric model}
\label{sec:nonparametric}

Rather than relying on a learnable parametric model in the synthesis pass (such as setting the amplitude of uniform noise), we route the standard synthetic output through a UNet. The output of this UNet is then added to the original input, effectively learning a residual that optimizes the synthetic data. This approach eliminates the need to manually define parametric models for each type of augmentation. Such a model would however be deterministic, whereas we aim to implement a randomization network. Inspired by diffusion networks, we concatenate the input image with a channel of Gaussian noise, thereby providing additional entropy to the network, leading to the following formulation:
\begin{align}
\boldsymbol{\xi} & \sim \mathcal{N}_{N}(0, 1), \\
\mathbf{x}_\mathrm{synth} & \leftarrow \mathbf{x}_\mathrm{synth} +
A_{\boldsymbol\theta}([\mathbf{x}_\mathrm{synth}, \boldsymbol{\xi}]) ~. 
\end{align}
Furthermore, in order to obviate forcing the augmentation network to learn an identity mapping, we apply the parametric noise augmentation described in~\Cref{eq:hypernoise} to its output. 

\section{Experiments}

Our experiments aim to answer two fundamental questions: \textbf{(1)} what training environment is deemed optimal by \emph{Learn2Synth} when the domain gap between $\mathbf{x}_\mathrm{real}$ and $\mathbf{x}_\mathrm{synth}$ is known? \textbf{(2)} can \emph{Learn2Synth} bridge the on-domain performance gap with supervised approaches without paying a penalty off-domain?  We tackle \textbf{(1)} in \Cref{sec:exp_syn} using synthetic experiments, in which $\mathbf{x}_\mathrm{real}$ is defined as a stochastic function $A_{\skew{2}\hat{\boldsymbol\theta}}(\mathbf{x}_\mathrm{synth})$ with known parameters ${\skew{2}\hat{\boldsymbol\theta}}$. The parameters of that same function $A_{\boldsymbol\theta}$ are then optimized using our framework, and the optimal parameters ${\boldsymbol\theta}^\ast$ are compared to the true parameters ${\skew{2}\hat{\boldsymbol\theta}}$. We tackle \textbf{(2)} in \Cref{sec:exp_auto} using two experiments with real MRI data, that mimic two typical training strategies for brain segmentation networks. In the first experiment, we use a single imaging contrast and a large set of ``soft truth'' labels obtained with an automated method. In the second experiment, we use a small number of high-quality ``ground truth'' labels and investigate generalizability across contrasts. Please find the implementation details in the supplementary material.


\subsection{Experiments on synthetic datasets}
\label{sec:exp_syn}

\myparagraph{Dataset.} We automatically segmented 434 images from the OASIS dataset~\cite{lamontagne2019oasis}, using FreeSurfer~\cite{fischl2002whole} and used the resulting label maps throughout. 

\myparagraph{Settings.} Here, synthetic images ($\mathbf{x}_\mathrm{synth}$) are noise-free, while ``real'' images ($\mathbf{x}_\mathrm{real}$) are obtained by adding noise and/or INU to noise-free images, with predefined parameter ${\skew{2}\hat{\boldsymbol\theta}}$. A \emph{Learn2Synth} model is then trained on this joint dataset, using either the parametric or nonparametric augmentation model $A_{\boldsymbol\theta}$. In each sub-experiment, we vary the value of the true parameter ${\skew{2}\hat{\boldsymbol\theta}}$ and compare it with the value of the learned parameter ${\boldsymbol\theta^\ast}$. In the parametric case, if ${\boldsymbol\theta^\ast}$ converged to  ${\skew{2}\hat{\boldsymbol\theta}}$, we would effectively obtain the same segmentation network as if $S_{\boldsymbol\phi}$ had been directly trained on $\mathbf{x}_\mathrm{real}$. However, there is no built-in incentive for \emph{Learn2Synth} to recover the exact same parameters. Instead, it is solely tasked with optimizing segmentation performance on $\mathbf{x}_\mathrm{real}$. 

\myparagraph{Baselines.} In each sub-experiment, we train two $\emph{SynthSeg}$ baselines: one where the training data uses the same augmentation parameters as the ``real'' data (Naive SynthSeg), and one where it uses the parameters inferred by \emph{Learn2Synth} (Optimized SynthSeg). In cases where the inferred parameters ${\boldsymbol\theta^\ast}$ differ from the ``real'' parameters  ${\skew{2}\hat{\boldsymbol\theta}}$, these baselines allow us to verify that the inferred parameters are indeed optimal.


\subsubsection{Gaussian noise}
\label{sec:noise_only}
We use the homogeneous noise model described in \Cref{eq:noise}, with different values of $\sigma = \{0, 0.05, 0.1, 0.15\}$, as well as a version where $\sigma$ is randomly sampled in $U[0.025, 0.2]$ at each step. In all cases, the learnable augmentation model $A_{\boldsymbol\theta}$ implements \Cref{eq:noise}. We report the inferred $\sigma$s for different noise-level settings in~\Cref{table:noise}. Inferred parameters match well with the preset parameters, although we note that the optimal solution could have been a different set of parameters in which the optimal training noise levels might be larger (or smaller) than those in the test images, should they yield higher real-world accuracy. 
\begin{table}[ht]
\centering
\scriptsize
\setlength{\tabcolsep}{8pt}
\begin{tabular}{cccccc}
\hline
Preset $\hat{\sigma}$ & 0 & 0.050 & 0.100 & 0.150 & [0.025 0.2] \\ 
\hline
Inferred $\sigma^\ast$ & 0.001 & 0.042 & 0.098 & 0.146 & 0.134 \\
\hline
\end{tabular}
\caption{Inferred noise $\sigma$ under the ``noise-only'' setting.}
\label{table:noise}
\end{table}


In~\Cref{table:seg_noise}, each row denotes the segmentation accuracy of models trained on different levels of noise when tested on images with some other level of noise. For example, the first block (\textbf{$\sigma=0$}) shows the accuracy of models trained on noise-free data when tested on images with five different noise levels. Conversely, each column contains the results of all models, when tested on images with a given amount of noise. 
We highlight the highest (bold) and second highest (underline) numbers in each column. 
As expected, performance declines as the noise level increases in the test set. Additionally, for images subjected to a specific noise level, the model trained under the same noise conditions achieves optimal performance. This is evidenced by the highlighted diagonal. This finding is in line with results from \Cref{table:noise}. Most importantly, when tested on noisy images under models trained with matching noise conditions, \emph{Learn2Synth} consistently outperforms naive SynthSeg. In theory, after convergence, \emph{Learn2Synth} and Optimized SynthSeg should yield identical results, which is also verified in practice.

\begin{table}[!h]
\centering
\scriptsize
\setlength{\tabcolsep}{5pt}
\begin{tabular}{ccccccc}
\hline
\diagbox{Train}{Test} & Model & $\sigma$ = 0 & $\sigma$ = 0.05 & $\sigma$ = 0.1 & $\sigma$ = 0.15 & $\sigma \sim U$ \\ \hline
\multirow{3}{*}{$\sigma$ = 0}     & N &  0.913 & 0.310 & 0.209 & 0.142 & 0.205\\
& \emph{L} & \textbf{0.923}    &  0.346    &  0.172    &  0.092    &  0.151 \\
& O & \underline{0.919} & 0.361 & 0.189 & 0.105 & 0.167\\
\hline
\multirow{3}{*}{$\sigma$ = 0.05}           & N &  0.889 & 0.874 &  0.718 & 0.520 & 0.646  \\
& \emph{L} & 0.900    &  \underline{0.887}    &  0.744    &  0.492    &  0.635 \\
& O & 0.898 & \textbf{0.890} & 0.736 & 0.508 & 0.651\\
\hline
\multirow{3}{*}{$\sigma$ = 0.1}            & N &     0.867 & 0.858 & 0.841 & 0.772 & 0.815\\
& \emph{L} & 0.874    &  0.873    &  \textbf{0.858}    &  0.799    &  0.831 \\
& O & 0.871 & 0.869 & \underline{0.847} & 0.810 & 0.829\\
\hline
\multirow{3}{*}{$\sigma$ = 0.15}           & N &  0.864 & 0.857 & 0.847 & 0.829 & 0.838   \\
& \emph{L} & 0.860    &  0.855    &  0.854    &  \textbf{0.840}    &  0.834 \\
& O & 0.867 & 0.860 & 0.852 & \underline{0.836} & 0.841 \\
\hline
\multirow{3}{*}{$\sigma \sim U$}  & N & 0.870 & 0.867 & 0.850 & 0.826 & 0.831    \\
& \emph{L} & 0.866    &  0.864    &  0.851    &  0.833    &  \underline{0.839} \\
& O & 0.872 & 0.870 & 0.853 & 0.837 & \textbf{0.843}\\
\hline
\end{tabular}
\caption{Segmentation accuracy (Dice) for naive SynthSeg (N), \emph{Learn2Synth} (\emph{L}) and Optimized SynthSeg (O).}
\label{table:seg_noise}
\end{table}

\subsubsection{Intensity non-uniformity} 
\label{sec:parametric_both}

\begin{figure}[t]
    \centering
     \setlength{\tabcolsep}{0.5pt}
    \begin{tabular}{ccc}
    \includegraphics[width=.16\textwidth]{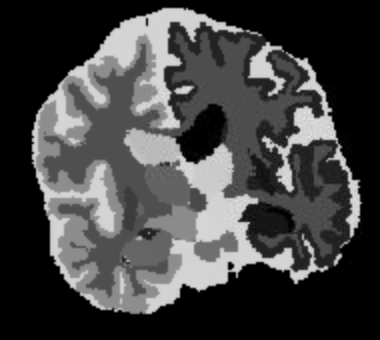} &
    \includegraphics[width=.16\textwidth]{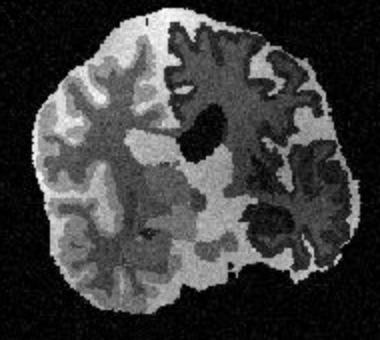} &
    \includegraphics[width=.16\textwidth]{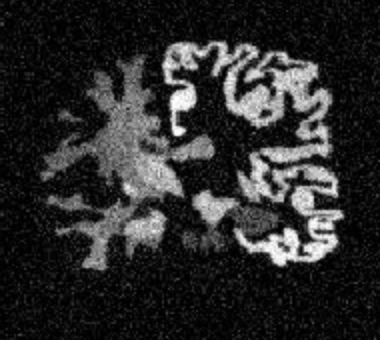}\\
     (a) Noise-free image & (b) Learned image & (c) Real image\\
    \end{tabular}
    \caption{Illustration of learned synthesized image. (a) Noise-free image, (b) learned synthesized image, and (c) real image.}
    \label{fig:parametric}
\end{figure}

Here, we use both the INU model described in \Cref{eq:bias} and the homogeneous noise model described in \Cref{eq:noise}. For the former, we use $K=3$ random fields with increasing spatial frequency ($\left\{M_1, M_2, M_3\right\} = \left\{2, 4, 8\right\}$). We denote the corresponding learnable parameters $c_\mathrm{low}$, $c_\mathrm{mid}$ and $c_\mathrm{high}$, with preset values $\hat{c}_\mathrm{low} = \hat{c}_\mathrm{mid} = \hat{c}_\mathrm{high} = 0.5$. Noise-free target ``real'', and learned images are shown in~\Cref{fig:parametric}. Inference results for the value of the noise parameter are reported in \Cref{table:middle_noise}, and are in line with those obtained in the noise-only experiment.

\begin{table}[ht]
\centering
\setlength{\tabcolsep}{8pt}
\scriptsize
\begin{tabular}{cccccc}
\hline
Preset $\hat\sigma$ & 0 & 0.050 & 0.100 & 0.150 & [0.025 0.2] \\ \hline
Inferred $\sigma^\ast$ & 0.018 & 0.049 & 0.085 & 0.147 & 0.134  \\
\hline
\end{tabular}
\caption{Inferred noise $\sigma$ under the ``noise and INU'' setting.}
\label{table:middle_noise}
\end{table}

Similar to the ``\textbf{parametric noise}'' setting, in addition to examining the inferred noise levels, we also report the segmentation accuracy for various noise-level settings in~\Cref{table:bias_middle_seg_learn2synth} for the \emph{Learn2Synth} model. We highlight the highest (bold) and second highest (underline) numbers in each column as well. Notably, both of the key observations from~\Cref{sec:noise_only} hold true for the scenario where both the noise and the bias field are modeled. This continued consistency demonstrates that our approach performs well under varying conditions, whether noise alone or both noise and bias fields are considered.

\begin{table}[ht]
\setlength{\tabcolsep}{3pt}
\centering
\scriptsize
\begin{tabular}{ccccccc}
\hline
Model & \diagbox{Train}{Test} & $\sigma$ = 0 &  $\sigma$ = 0.05 
& $\sigma$ = 0.1 & $\sigma$ = 0.15 & $\sigma \sim U$ \\ \hline
\multirow{5}{*}{Parametric} 
& $\sigma$ = 0  &  \textbf{0.924} &  0.867 &  0.748 &  0.619 &  0.729    \\
& $\sigma$ = 0.05        &  0.900 &  \textbf{0.894} &  0.867 &  0.696 &  0.744    \\
& $\sigma$ = 0.1         &  0.894 &  0.889 &  \textbf{0.872} &  0.832 &  0.849    \\
& $\sigma$ = 0.15        &  0.881 &  0.869 & 0.856 &  \textbf{0.854} &  0.862    \\
& $\sigma \sim U$       &  0.875 &  0.868 &  0.870 &  0.851 &  \textbf{0.867}    \\
\hline
\multirow{5}{*}{Nonparametric} 
& $\sigma$ = 0  & \underline{0.903} & 0.871 & 0.752 & 0.603 & 0.769 \\
& $\sigma$ = 0.05        & 0.891 & \underline{0.886} & 0.859 & 0.749 & 0.792 \\
& $\sigma$ = 0.1         &  0.884 & 0.877 & \underline{0.865} & 0.828 & 0.839 \\
& $\sigma$ = 0.15        & 0.878  & 0.864 & 0.851 & \underline{0.837} & 0.844 \\
& $\sigma \sim U$       & 0.869 & 0.858 & 0.843 & 0.831 & \underline{0.850} \\
\hline
\end{tabular}
\caption{Segmentation accuracy for \emph{Learn2Synth} under the parametric ``noise and INU'' and nonparametric settings.}
\label{table:bias_middle_seg_learn2synth}
\end{table}

\begin{figure*}[!ht]
    \centering
     \setlength{\tabcolsep}{0.5pt}
    \begin{tabular}{cccccc}
    \includegraphics[width=.16\textwidth]{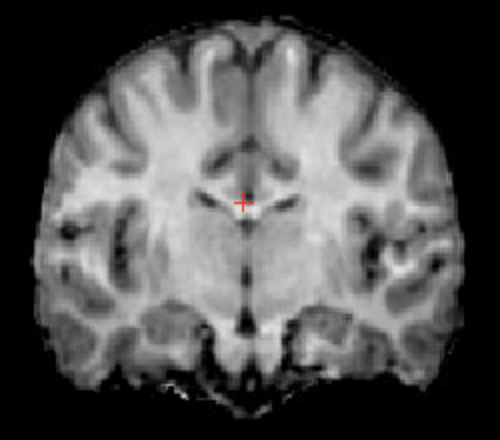} &
    \includegraphics[width=.16\textwidth]{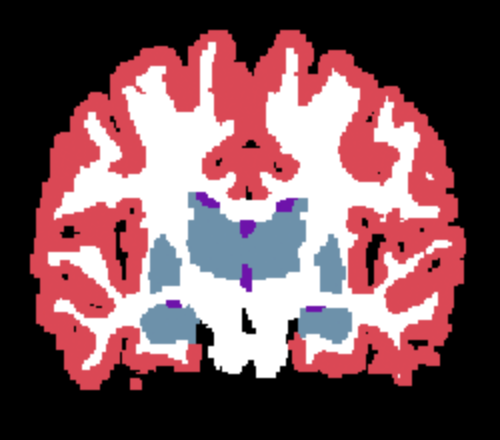} &
    \includegraphics[width=.16\textwidth]{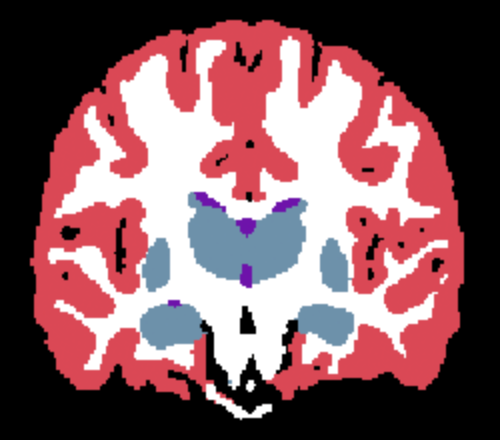} &
    \includegraphics[width=.16\textwidth]{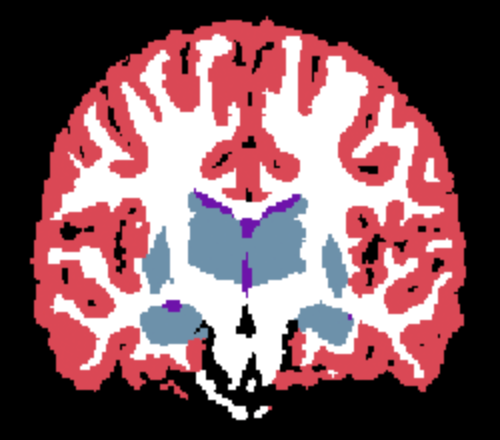} &   \includegraphics[width=.16\textwidth]{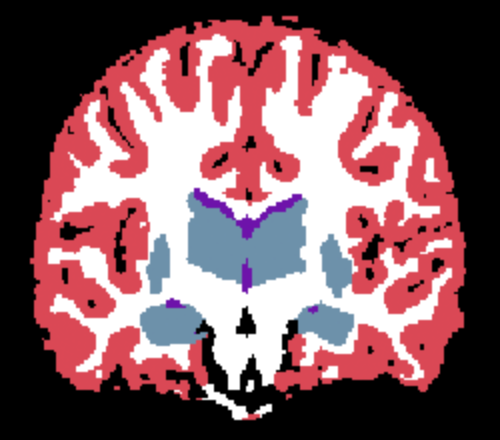} &
    \includegraphics[width=.16\textwidth]{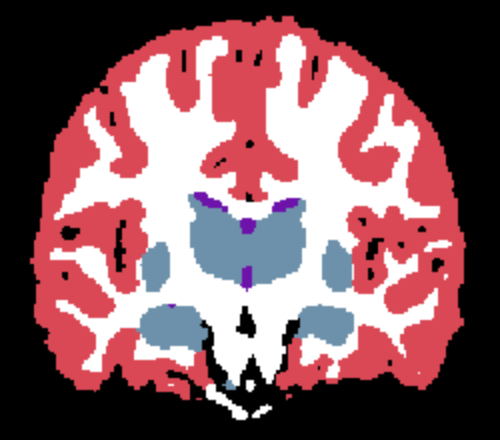} \\

    \includegraphics[width=.16\textwidth]{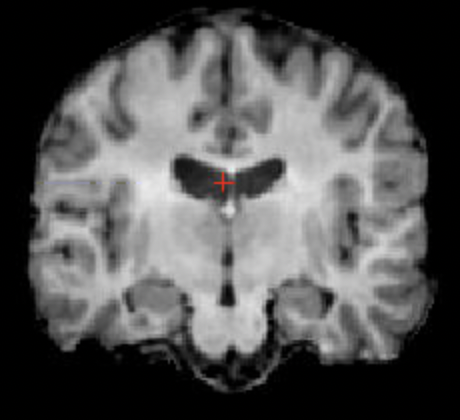} &
    \includegraphics[width=.16\textwidth]{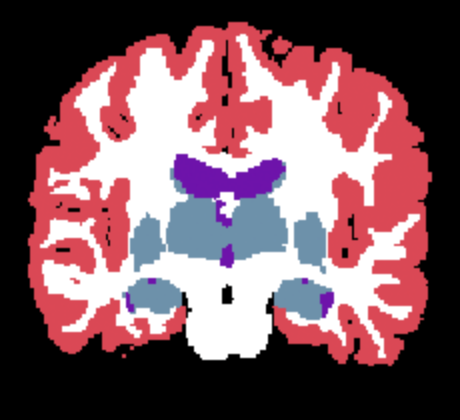} &
    \includegraphics[width=.16\textwidth]{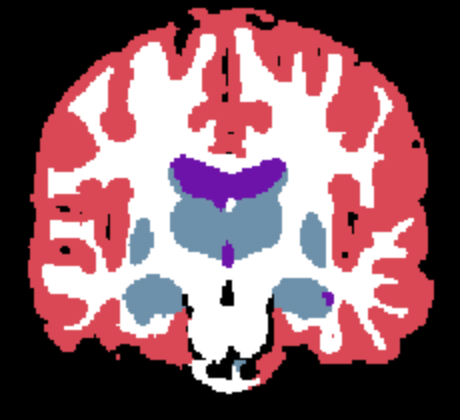} &
    \includegraphics[width=.16\textwidth]{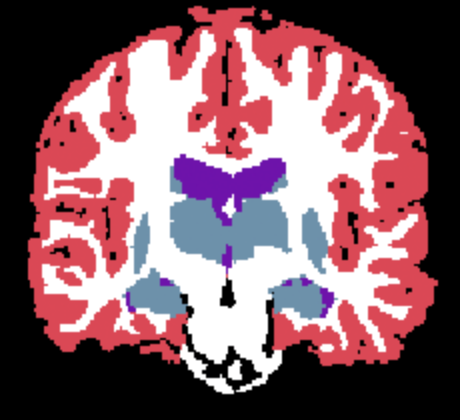} &   \includegraphics[width=.16\textwidth]{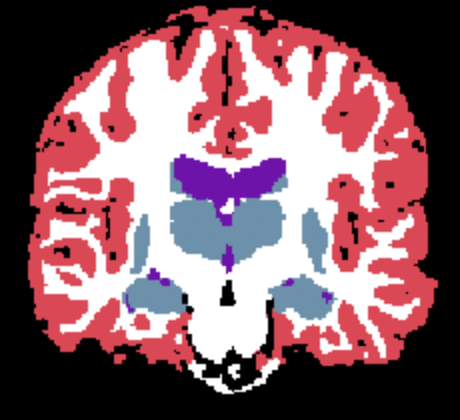} &
    \includegraphics[width=.16\textwidth]{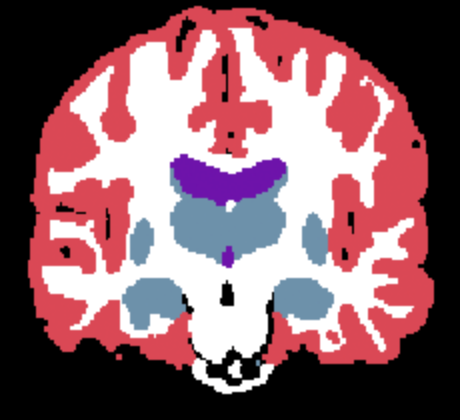}\\
     (a) Input & (b) Ground Truth & (c) Supervised & (d) SAMSEG & (e) Naive SynthSeg & (f) \emph{Learn2Synth} \\
    \end{tabular}
    \caption{Qualitative results on OASIS3 and ABIDE. (a), (b) show the input and segmentation. (c)-(f) show the segmentations of different methods.}
    \label{fig:qualitative}
\end{figure*}

We have also trained \Cref{eq:hypernoise}'s ``nonparametric + noise'' model, instead of a ``INU + noise'' model, on the same training data (\emph{i.e.}, generated using the parametric INU + noise function). While inference results are in line with previous sections (not shown here), segmentation accuracy is slightly lower than that obtained with an accurate parametric model. This highlights that \emph{if} a low-dimensional parametric model of the data is known, it should be used over a high-dimensional nonparametric alternative.




\subsection{Experiments on real-world datasets}
\label{sec:exp_auto}

In the previous section, we have shown that using synthetic data \emph{Learn2Synth} is able to transform its synthetic training domain so that it matches its ``real'' training domain. Furthermore, when domains are not matched exactly, the resulting network is more accurate than a network trained on the ``real'' domain. However, the ``real'' domain that we used was far from realistic, and the reality gap between typical synthetic data and real MR images is much wider. In this section, we aim to demonstrate the performance of \emph{Learn2Synth} in real world scenarios.

\myparagraph{Our first experiment} (\Cref{sec:exp_freesurfer}) aims to show that even a well-specified synthesis model can be improved if real data is available. We use images from two large multi-site datasets, along with ``soft truth'' labels obtained with a standard automated method.

\myparagraph{Our second experiment} (\Cref{sec:exp_manual}) tackles a more interesting problem, where a large set of ``soft truth'' labels is available, but only a handful of manual ``ground truth'' labels are. In this context, we further investigate the performance of our method and its baselines on a held-out test set whose contrast differs from the one used during training.

\subsubsection{\emph{Learn2Synth} outperforms mixed training}
\label{sec:exp_freesurfer}

\myparagraph{Datasets.} We automatically segmented anatomical images from the \textbf{ABIDE}~\cite{di2014autism} and \textbf{OASIS3}~\cite{lamontagne2019oasis} datasets, using FreeSurfer~\cite{fischl2002whole} and merged the resulting segmentation maps into four labels (white matter, cortical gray matter, subcortical gray matter, and ventricles). 

\myparagraph{Baselines.} We compare \emph{Learn2Synth} with a simple \textbf{Supervised UNet}~\cite{ronneberger2015u} trained on real data; \textbf{SAMSEG}~\cite{puonti2016fast}, a state-of-the-art unsupervised Bayesian segmentation framework; and \textbf{SynthSeg}~\cite{billot2023synthseg}, a state-of-the-art domain randomization approach for brain image segmentation. Note that since we are running our experiments in 2D, we do not use the published SynthSeg model, but instead train, from scratch, our own version with matched hyper-parameters. We call this baseline \textbf{Naive SynthSeg} for clarity. We also train another SynthSeg model on a dataset augmented with a small number of real images (\textbf{Mixed SynthSeg}), and we further finetune Naive SynthSeg on the same small number of real images (\textbf{Finetuned SynthSeg}). Additionally, we add another relevant adversarial-based baseline \textit{AdvChain}~\cite{chen2022enhancing}.

\setlength{\tabcolsep}{7pt}
\begin{table}[ht]
\centering
\scriptsize
\begin{tabular}{ccc}
\hline
Method & ABIDE & OASIS3 \\ \hline
Supervised UNet & 0.908 & 0.899\\
\hline
SAMSEG    & 0.875 & 0.841\\
\hline
Naive SynthSeg & 0.869 & 0.831\\
Mixed SynthSeg    & 0.875 & 0.854 \\
Finetuned SynthSeg  & 0.871  & 0.847 \\
\textit{AdvChain} & 0.867 & 0.848 \\
\hline
\emph{Learn2Synth} (parametric setting fixed $\sigma$)        & 0.878 & 0.860\\
\emph{Learn2Synth} (parametric setting varying $\sigma$)          &  \textbf{0.881} & 0.857\\
\emph{Learn2Synth} (nonparametric setting fixed $\sigma$)        & 0.879 & 0.875\\
\emph{Learn2Synth} (nonparametric setting varying $\sigma$)       &  0.874 & \textbf{0.881}\\
\hline
\end{tabular}
\caption{Segmentation accuracy for \emph{Learn2Synth} and the baselines on ABIDE and OASIS3.}
\label{table:real}
\end{table}

\begin{figure}[ht]
    \centering
    \includegraphics[width=\columnwidth]{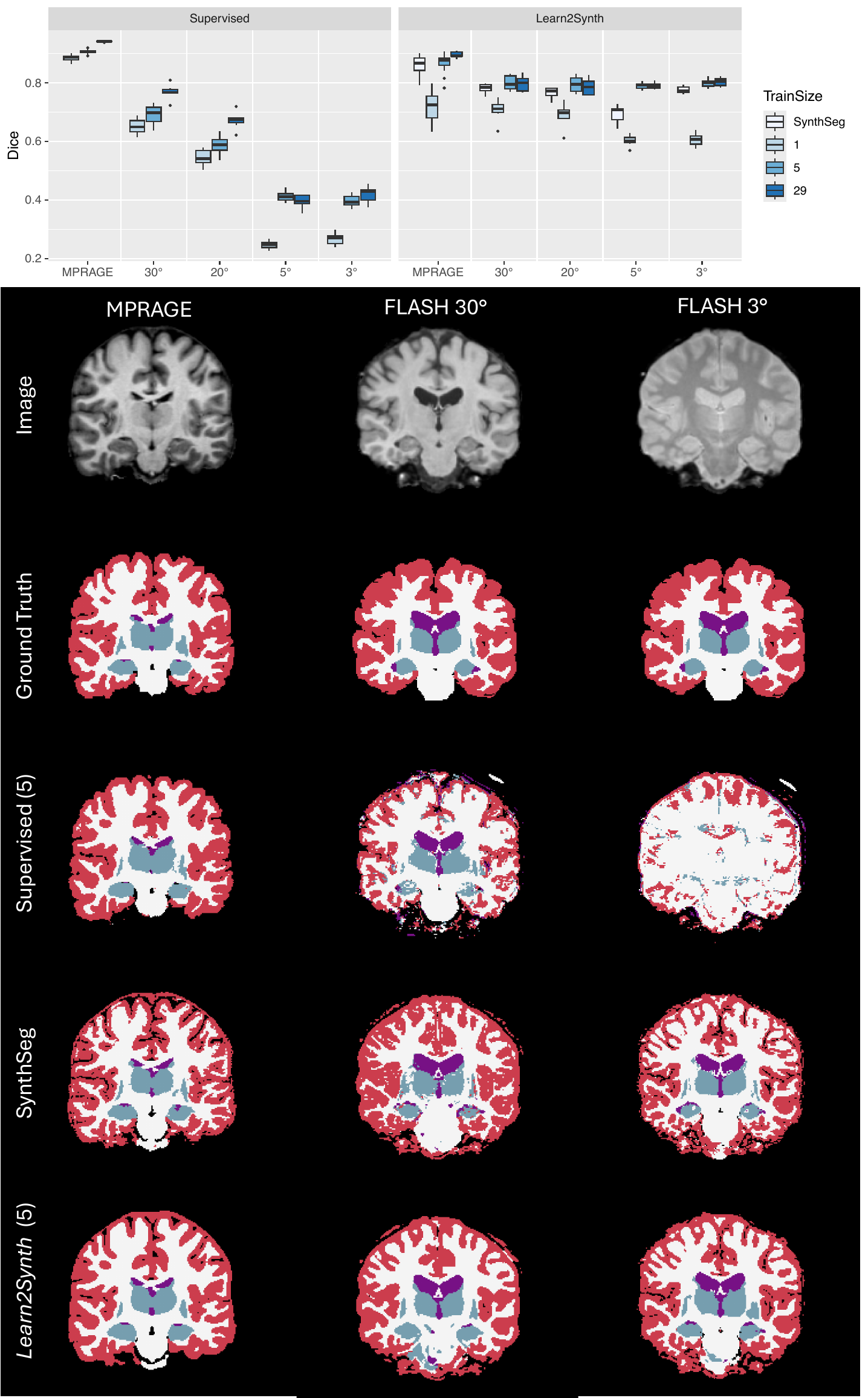}
    \caption{Results obtained when training on MPRAGE scans and testing on FLASH scans.}
    \label{fig:flash}
\end{figure}

\begin{table*}[!ht]
\setlength{\tabcolsep}{8pt}
    \centering
    \scriptsize
    \begin{tabular}{l c c c c c c}
        \hline
        Setting & \# of Train, Val., Test & Test Set (MPRAGE) & $3^{\circ}$ & $5^{\circ}$ & $20^{\circ}$ & $30^{\circ}$ \\
        \hline
        SynthSeg & / & $0.861\pm0.028$ & $0.776\pm0.012$ & $0.694\pm0.027$ & $0.766\pm0.018$ & $0.781\pm0.016$ \\
        \hline
        Supervised UNet & 29, 5, 5 & $\mathbf{0.941}\pm 0.002$ & $0.419\pm0.025$ & $0.396\pm0.020$ & $0.671\pm0.026$ & $0.769\pm0.022$ \\
        Supervised UNet & 5, 5, 29 & $0.907\pm0.007$ & $0.397\pm0.018$ & $0.413\pm0.016$ & $0.586\pm0.033$ & $0.692\pm0.033$ \\
        Supervised UNet & 1, 5, 33 & $0.885\pm0.010$ & $0.267\pm0.019$ & $0.247\pm0.013$ & $0.544\pm0.026$ & $0.651\pm0.025$ \\
        \hline
        \emph{Learn2Synth} & 29, 5, 5 & $0.895\pm0.011$ & $\mathbf{0.804}\pm0.014$ & $\mathbf{0.789}\pm0.010$ & $0.785\pm0.025$ & $0.797\pm0.023$ \\
        \emph{Learn2Synth} & 5, 5, 29 & $0.867\pm0.030$ & $0.798\pm0.013$ & $\mathbf{0.789}\pm0.010$ & $\mathbf{0.795}\pm0.026$ & $\mathbf{0.799}\pm0.024$ \\
        \emph{Learn2Synth} & 1, 5, 33 & $0.718\pm0.045$ & $0.606\pm0.020$ & $0.603\pm0.017$ & $0.690\pm0.036$ & $0.707\pm0.032$ \\
        \hline
    \end{tabular}%
        \caption{Performance comparison of different models across various datasets.}
    \label{tab:manual_results}
\end{table*}
\myparagraph{Results.} Qualitative segmentation results are presented in~\Cref{fig:qualitative} and quantitative results are summarized in \Cref{table:real}.~\Cref{fig:qualitative} shows that \emph{Learn2Synth} produces cleaned and more accurate segmentation results compared to SAMSEG and naive SynthSeg, which is further validated by the higher Dice scores reported in~\Cref{table:real}. \emph{Learn2Synth} also obtains higher Dice scores than the two SynthSeg variants that incorporate real data. These results highlight the effectiveness of \emph{Learn2Synth} in enhancing segmentation quality for complex real-world data.

\subsubsection{\emph{Learn2Synth} generalizes to unseen contrasts}
\label{sec:exp_manual}

\myparagraph{Datasets.} We further validate the proposed algorithm on \textbf{Buckner39}~\cite{fischl2002whole}, a dataset with 39 manually labeled MPRAGE images, and the \textbf{Freesurfer maintenance dataset}~\cite{fischl2004sequence}, which contains images acquired in 8 subjects with a FLASH sequence at multiple flip angles. In contrast with MPRAGE, the FLASH sequence does not apply an inversion pulse, which results in lower cortical contrast. Furthermore, different flip angles give rise to different contrasts. 

\myparagraph{Settings.} We trained a \textbf{SynthSeg} baseline using the same settings as in the previous section. We also trained three \textbf{supervised UNets} and three \emph{Learn2Synth} models on the Buckner39 dataset, with different training set sizes. Varying the training size allows different scenarios of data availability to be modeled. All models were tested on the remaining test portion of Buckner39 and on all images from the Freesurfer maintenance dataset. Models that obtained the best performance on their validation set were selected for evaluation on the test set. To quantify the impact of this validation step, we also provide results obtained with fully converged models in the supplementary material.

\myparagraph{Results.} Quantitative results are summarized in~\Cref{tab:manual_results}, and graphically represented in~\Cref{fig:flash}. The supervised UNet achieves the highest Dice score on Buckner39, particularly when trained with 29 training samples, reaching $0.941 \pm 0.002$, which is not surprising. However, its performance drops significantly on FLASH scans, highlighting its limited generalization ability. \emph{Learn2Synth}, on the other hand, demonstrates improved generalization across datasets, achieving Dice scores of approximately 0.79 across all FLASH sequences when trained with as few as 5 training samples. Surprisingly, \emph{Learn2Synth} improves its Dice over the SynthSeg baseline on MPRAGE scans (as expected) but also on all FLASH scans. These results suggest that \emph{Learn2Synth} effectively generalizes to unseen imaging conditions.



\section{Conclusion}
In conclusion, we present \emph{Learn2Synth}, a novel approach to enhance segmentation performance without directly exposing the segmentation network to real data. 
Our experimental results on both synthetic and real datasets confirm that \emph{Learn2Synth} reaches substantially higher segmentation performance accuracy than naive SynthSeg, and generalizes considerably better to unseen contrasts than supervised models as well as SynthSeg. 

\myparagraph{Acknowledgements.} YB is supported by a fellowship from the Royal Society (NIF$\verb|\|$R1$\verb|\|$232460). XZ was supported by a postdoctoral fellowship from Huntington's Disease Society of America human biology project. Support for this research was provided in part by the BRAIN Initiative Cell Atlas Network (BICAN) grants U01MH117023, UM1MH134812 and UM1MH130981, the
Brain Initiative Brain Connects consortium (U01NS132181, 1UM1NS132358-01), the National Institute for Biomedical Imaging and Bioengineering (1R01EB023281, R21EB018907, R01EB019956, P41EB030006, 1R01EB031114), the National Institute on Aging (R21AG082082, 1R01AG064027, R01AG016495, 1R01AG070988, 1RF1AG080371, 1R21NS138995), the National Institute of Mental Health (R01 MH123195, R01 MH121885, 1RF1MH123195), the National Institute for Neurological Disorders and Stroke, (1U24NS135561-01, R01NS070963, 2R01NS083534, R01NS105820, R25NS125599), and was made possible by the resources provided by Shared Instrumentation Grants 1S10RR023401, 1S10RR019307, and 1S10RR023043. Additional support was provided by the NIH Blueprint for Neuroscience Research (5U01-MH093765), part of the multi-institutional Human Connectome Project. Much of the computation resources required for this research was performed on computational hardware generously provided by the Massachusetts Life Sciences Center (https://www.masslifesciences.com/). In addition, BF is an advisor to DeepHealth, a company whose medical pursuits focus on medical imaging and measurement technologies. BF's interests were reviewed and are managed by Massachusetts General Hospital and Partners HealthCare in accordance with their conflict of interest policies.

{
    \small
    \bibliographystyle{ieeenat_fullname}
    \bibliography{main}

\begin{thebibliography}{90}
\providecommand{\natexlab}[1]{#1}
\providecommand{\url}[1]{\texttt{#1}}
\expandafter\ifx\csname urlstyle\endcsname\relax
  \providecommand{\doi}[1]{doi: #1}\else
  \providecommand{\doi}{doi: \begingroup \urlstyle{rm}\Url}\fi

\bibitem[Al~Olaimat et~al.(2020)Al~Olaimat, Lee, Kim, Kim, and Kim]{al2020learning}
Mohammad Al~Olaimat, Dongeun Lee, Youngsoo Kim, Jonghyun Kim, and Jinoh Kim.
\newblock A learning-based data augmentation for network anomaly detection.
\newblock In \emph{ICCCN}, 2020.

\bibitem[Antonelli et~al.(2022)Antonelli, Reinke, Bakas, Farahani, Kopp-Schneider, Landman, Litjens, Menze, Ronneberger, Summers, et~al.]{antonelli2022medical}
Michela Antonelli, Annika Reinke, Spyridon Bakas, Keyvan Farahani, Annette Kopp-Schneider, Bennett~A Landman, Geert Litjens, Bjoern Menze, Olaf Ronneberger, Ronald~M Summers, et~al.
\newblock The medical segmentation decathlon.
\newblock \emph{Nature communications}, 2022.

\bibitem[Baldwin(2022)]{baldwin2022deep}
Breck Baldwin.
\newblock Deep learning does not replace bayesian modeling: Comparing research use via citation counting.
\newblock \emph{Applied AI Letters}, 2022.

\bibitem[Baydin et~al.(2018)Baydin, Cornish, Mart{\'{\i}}nez{-}Rubio, Schmidt, and Wood]{baydin2017online}
Atilim~Gunes Baydin, Robert Cornish, David Mart{\'{\i}}nez{-}Rubio, Mark Schmidt, and Frank Wood.
\newblock Online learning rate adaptation with hypergradient descent.
\newblock In \emph{ICLR}, 2018.

\bibitem[Bengio et~al.(2011)Bengio, Bastien, Bergeron, Boulanger-Lewandowski, Breuel, Chherawala, Cisse, C{\^o}t{\'e}, Erhan, Eustache, et~al.]{bengio2011deep}
Yoshua Bengio, Fr{\'e}d{\'e}ric Bastien, Arnaud Bergeron, Nicolas Boulanger-Lewandowski, Thomas Breuel, Youssouf Chherawala, Moustapha Cisse, Myriam C{\^o}t{\'e}, Dumitru Erhan, Jeremy Eustache, et~al.
\newblock Deep learners benefit more from out-of-distribution examples.
\newblock In \emph{AISTATS}, 2011.

\bibitem[Billot et~al.(2023{\natexlab{a}})Billot, Greve, Puonti, Thielscher, Van~Leemput, Fischl, Dalca, Iglesias, et~al.]{billot2023synthseg}
Benjamin Billot, Douglas~N Greve, Oula Puonti, Axel Thielscher, Koen Van~Leemput, Bruce Fischl, Adrian~V Dalca, Juan~Eugenio Iglesias, et~al.
\newblock {SynthSeg}: Segmentation of brain {MRI} scans of any contrast and resolution without retraining.
\newblock \emph{MedIA}, 2023{\natexlab{a}}.

\bibitem[Billot et~al.(2023{\natexlab{b}})Billot, Magdamo, Cheng, Arnold, Das, and Iglesias]{billot2023robust}
Benjamin Billot, Colin Magdamo, You Cheng, Steven~E Arnold, Sudeshna Das, and Juan~Eugenio Iglesias.
\newblock Robust machine learning segmentation for large-scale analysis of heterogeneous clinical brain {MRI} datasets.
\newblock \emph{PNAS}, 2023{\natexlab{b}}.

\bibitem[Bluemke et~al.(2020)Bluemke, Moy, Bredella, Ertl-Wagner, Fowler, Goh, Halpern, Hess, Schiebler, and Weiss]{bluemke2020assessing}
David~A Bluemke, Linda Moy, Miriam~A Bredella, Birgit~B Ertl-Wagner, Kathryn~J Fowler, Vicky~J Goh, Elkan~F Halpern, Christopher~P Hess, Mark~L Schiebler, and Clifford~R Weiss.
\newblock Assessing radiology research on artificial intelligence: a brief guide for authors, reviewers, and readers—from the radiology editorial board, 2020.

\bibitem[Chaitanya et~al.(2019)Chaitanya, Karani, Baumgartner, Becker, Donati, and Konukoglu]{chaitanya2019semi}
Krishna Chaitanya, Neerav Karani, Christian~F Baumgartner, Anton Becker, Olivio Donati, and Ender Konukoglu.
\newblock Semi-supervised and task-driven data augmentation.
\newblock In \emph{IPMI}, 2019.

\bibitem[Chaitanya et~al.(2020)Chaitanya, Erdil, Karani, and Konukoglu]{chaitanya2020contrastive}
Krishna Chaitanya, Ertunc Erdil, Neerav Karani, and Ender Konukoglu.
\newblock Contrastive learning of global and local features for medical image segmentation with limited annotations.
\newblock In \emph{NeurIPS}, 2020.

\bibitem[Chandra et~al.(2022)Chandra, Xie, Ragan-Kelley, and Meijer]{chandra2022gradient}
Kartik Chandra, Audrey Xie, Jonathan Ragan-Kelley, and Erik Meijer.
\newblock Gradient descent: The ultimate optimizer.
\newblock In \emph{NeurIPS}, 2022.

\bibitem[Chartsias et~al.(2017)Chartsias, Joyce, Giuffrida, and Tsaftaris]{chartsias2017multimodal}
Agisilaos Chartsias, Thomas Joyce, Mario~Valerio Giuffrida, and Sotirios~A Tsaftaris.
\newblock Multimodal {MR} synthesis via modality-invariant latent representation.
\newblock \emph{TMI}, 2017.

\bibitem[Chen et~al.(2019)Chen, Dou, Chen, Qin, and Heng]{chen2019synergistic}
Cheng Chen, Qi Dou, Hao Chen, Jing Qin, and Pheng-Ann Heng.
\newblock Synergistic image and feature adaptation: Towards cross-modality domain adaptation for medical image segmentation.
\newblock In \emph{AAAI}, 2019.

\bibitem[Chen et~al.(2022)Chen, Qin, Ouyang, Li, Wang, Qiu, Chen, Tarroni, Bai, and Rueckert]{chen2022enhancing}
Chen Chen, Chen Qin, Cheng Ouyang, Zeju Li, Shuo Wang, Huaqi Qiu, Liang Chen, Giacomo Tarroni, Wenjia Bai, and Daniel Rueckert.
\newblock Enhancing {MR} image segmentation with realistic adversarial data augmentation.
\newblock \emph{MedIA}, 2022.

\bibitem[Chen et~al.(2021)Chen, Lu, Yu, Luo, Adeli, Wang, Lu, Yuille, and Zhou]{chen2021transunet}
Jieneng Chen, Yongyi Lu, Qihang Yu, Xiangde Luo, Ehsan Adeli, Yan Wang, Le Lu, Alan~L Yuille, and Yuyin Zhou.
\newblock Transunet: Transformers make strong encoders for medical image segmentation.
\newblock \emph{arXiv preprint arXiv:2102.04306}, 2021.

\bibitem[Chen et~al.(2014)Chen, Papandreou, Kokkinos, Murphy, and Yuille]{chen2014semantic}
Liang-Chieh Chen, George Papandreou, Iasonas Kokkinos, Kevin Murphy, and Alan~L Yuille.
\newblock Semantic image segmentation with deep convolutional nets and fully connected {CRFs}.
\newblock \emph{arXiv preprint arXiv:1412.7062}, 2014.

\bibitem[Chen et~al.(2017)Chen, Papandreou, Schroff, and Adam]{chen2017rethinking}
Liang-Chieh Chen, George Papandreou, Florian Schroff, and Hartwig Adam.
\newblock Rethinking atrous convolution for semantic image segmentation.
\newblock \emph{arXiv preprint arXiv:1706.05587}, 2017.

\bibitem[Chen et~al.(2018)Chen, Papandreou, Kokkinos, Murphy, and Yuille]{chen2018deeplab}
Liang-Chieh Chen, George Papandreou, Iasonas Kokkinos, Kevin Murphy, and Alan~L Yuille.
\newblock Deeplab: Semantic image segmentation with deep convolutional nets, atrous convolution, and fully connected {CRFs}.
\newblock \emph{TPAMI}, 2018.

\bibitem[Chen et~al.(2020)Chen, Kornblith, Norouzi, and Hinton]{chen2020simple}
Ting Chen, Simon Kornblith, Mohammad Norouzi, and Geoffrey Hinton.
\newblock A simple framework for contrastive learning of visual representations.
\newblock In \emph{ICML}, 2020.

\bibitem[Cubuk et~al.(2019)Cubuk, Zoph, Mane, Vasudevan, and Le]{cubuk2019autoaugment}
Ekin~D Cubuk, Barret Zoph, Dandelion Mane, Vijay Vasudevan, and Quoc~V Le.
\newblock Autoaugment: Learning augmentation strategies from data.
\newblock In \emph{CVPR}, 2019.

\bibitem[Di~Martino et~al.(2014)Di~Martino, Yan, Li, Denio, Castellanos, Alaerts, Anderson, Assaf, Bookheimer, Dapretto, et~al.]{di2014autism}
Adriana Di~Martino, Chao-Gan Yan, Qingyang Li, Erin Denio, Francisco~X Castellanos, Kaat Alaerts, Jeffrey~S Anderson, Michal Assaf, Susan~Y Bookheimer, Mirella Dapretto, et~al.
\newblock The autism brain imaging data exchange: towards a large-scale evaluation of the intrinsic brain architecture in autism.
\newblock \emph{Molecular psychiatry}, 2014.

\bibitem[Dou et~al.(2019)Dou, Ouyang, Chen, Chen, Glocker, Zhuang, and Heng]{dou2019pnp}
Qi Dou, Cheng Ouyang, Cheng Chen, Hao Chen, Ben Glocker, Xiahai Zhuang, and Pheng-Ann Heng.
\newblock {PnP-AdaNet}: Plug-and-play adversarial domain adaptation network at unpaired cross-modality cardiac segmentation.
\newblock \emph{IEEE Access}, 2019.

\bibitem[Fischl et~al.(2002)Fischl, Salat, Busa, Albert, Dieterich, Haselgrove, Van Der~Kouwe, Killiany, Kennedy, Klaveness, et~al.]{fischl2002whole}
Bruce Fischl, David~H Salat, Evelina Busa, Marilyn Albert, Megan Dieterich, Christian Haselgrove, Andre Van Der~Kouwe, Ron Killiany, David Kennedy, Shuna Klaveness, et~al.
\newblock Whole brain segmentation: automated labeling of neuroanatomical structures in the human brain.
\newblock \emph{Neuron}, 2002.

\bibitem[Fischl et~al.(2004)Fischl, Salat, Van Der~Kouwe, Makris, S{\'e}gonne, Quinn, and Dale]{fischl2004sequence}
Bruce Fischl, David~H Salat, Andr{\'e}~JW Van Der~Kouwe, Nikos Makris, Florent S{\'e}gonne, Brian~T Quinn, and Anders~M Dale.
\newblock Sequence-independent segmentation of magnetic resonance images.
\newblock \emph{Neuroimage}, 2004.

\bibitem[Franceschi et~al.(2017)Franceschi, Donini, Frasconi, and Pontil]{franceschi2017forward}
Luca Franceschi, Michele Donini, Paolo Frasconi, and Massimiliano Pontil.
\newblock Forward and reverse gradient-based hyperparameter optimization.
\newblock In \emph{ICML}, 2017.

\bibitem[Frid-Adar et~al.(2018)Frid-Adar, Klang, Amitai, Goldberger, and Greenspan]{frid2018synthetic}
Maayan Frid-Adar, Eyal Klang, Michal Amitai, Jacob Goldberger, and Hayit Greenspan.
\newblock Synthetic data augmentation using {GAN} for improved liver lesion classification.
\newblock In \emph{ISBI}, 2018.

\bibitem[Ganin et~al.(2016)Ganin, Ustinova, Ajakan, Germain, Larochelle, Laviolette, March, and Lempitsky]{ganin2016domain}
Yaroslav Ganin, Evgeniya Ustinova, Hana Ajakan, Pascal Germain, Hugo Larochelle, Fran{\c{c}}ois Laviolette, Mario March, and Victor Lempitsky.
\newblock Domain-adversarial training of neural networks.
\newblock \emph{JMLR}, 2016.

\bibitem[Ganin et~al.(2018)Ganin, Kulkarni, Babuschkin, Eslami, and Vinyals]{ganin2018synthesizing}
Yaroslav Ganin, Tejas Kulkarni, Igor Babuschkin, SM~Ali Eslami, and Oriol Vinyals.
\newblock Synthesizing programs for images using reinforced adversarial learning.
\newblock In \emph{ICML}, 2018.

\bibitem[Hataya et~al.(2020)Hataya, Zdenek, Yoshizoe, and Nakayama]{hataya2020faster}
Ryuichiro Hataya, Jan Zdenek, Kazuki Yoshizoe, and Hideki Nakayama.
\newblock Faster autoaugment: Learning augmentation strategies using backpropagation.
\newblock In \emph{ECCV}, 2020.

\bibitem[He et~al.(2016)He, Zhang, Ren, and Sun]{he2016deep}
Kaiming He, Xiangyu Zhang, Shaoqing Ren, and Jian Sun.
\newblock Deep residual learning for image recognition.
\newblock In \emph{CVPR}, 2016.

\bibitem[He et~al.(2020)He, Fan, Wu, Xie, and Girshick]{he2020momentum}
Kaiming He, Haoqi Fan, Yuxin Wu, Saining Xie, and Ross Girshick.
\newblock Momentum contrast for unsupervised visual representation learning.
\newblock In \emph{CVPR}, 2020.

\bibitem[He et~al.(2021)He, Carass, Zuo, Dewey, and Prince]{he2021autoencoder}
Yufan He, Aaron Carass, Lianrui Zuo, Blake~E Dewey, and Jerry~L Prince.
\newblock Autoencoder based self-supervised test-time adaptation for medical image analysis.
\newblock \emph{MedIA}, 2021.

\bibitem[Ho et~al.(2020)Ho, Jain, and Abbeel]{ho2020denoising}
Jonathan Ho, Ajay Jain, and Pieter Abbeel.
\newblock Denoising diffusion probabilistic models.
\newblock In \emph{NeurIPS}, 2020.

\bibitem[Hoffman et~al.(2018)Hoffman, Tzeng, Park, Zhu, Isola, Saenko, Efros, and Darrell]{hoffman2018cycada}
Judy Hoffman, Eric Tzeng, Taesung Park, Jun-Yan Zhu, Phillip Isola, Kate Saenko, Alexei Efros, and Trevor Darrell.
\newblock {CyCADA}: Cycle-consistent adversarial domain adaptation.
\newblock In \emph{ICML}, 2018.

\bibitem[Hoffmann et~al.(2021)Hoffmann, Billot, Greve, Iglesias, Fischl, and Dalca]{hoffmann2021synthmorph}
Malte Hoffmann, Benjamin Billot, Douglas~N Greve, Juan~Eugenio Iglesias, Bruce Fischl, and Adrian~V Dalca.
\newblock {SynthMorph}: learning contrast-invariant registration without acquired images.
\newblock \emph{TMI}, 2021.

\bibitem[Hoopes et~al.(2022)Hoopes, Mora, Dalca, Fischl, and Hoffmann]{hoopes2022synthstrip}
Andrew Hoopes, Jocelyn~S Mora, Adrian~V Dalca, Bruce Fischl, and Malte Hoffmann.
\newblock {SynthStrip}: skull-stripping for any brain image.
\newblock \emph{NeuroImage}, 2022.

\bibitem[Huo et~al.(2018)Huo, Xu, Moon, Bao, Assad, Moyo, Savona, Abramson, and Landman]{huo2018synseg}
Yuankai Huo, Zhoubing Xu, Hyeonsoo Moon, Shunxing Bao, Albert Assad, Tamara~K Moyo, Michael~R Savona, Richard~G Abramson, and Bennett~A Landman.
\newblock Synseg-net: Synthetic segmentation without target modality ground truth.
\newblock \emph{TMI}, 2018.

\bibitem[Iglesias et~al.(2023)Iglesias, Billot, Balbastre, Magdamo, Arnold, Das, Edlow, Alexander, Golland, and Fischl]{iglesias2023synthsr}
Juan~E Iglesias, Benjamin Billot, Ya{\"e}l Balbastre, Colin Magdamo, Steven~E Arnold, Sudeshna Das, Brian~L Edlow, Daniel~C Alexander, Polina Golland, and Bruce Fischl.
\newblock {SynthSR}: A public {AI} tool to turn heterogeneous clinical brain scans into high-resolution t1-weighted images for {3D} morphometry.
\newblock \emph{Science advances}, 2023.

\bibitem[Isensee et~al.(2021)Isensee, Jaeger, Kohl, Petersen, and Maier-Hein]{isensee2021nnu}
Fabian Isensee, Paul~F Jaeger, Simon~AA Kohl, Jens Petersen, and Klaus~H Maier-Hein.
\newblock {nnU-Net}: a self-configuring method for deep learning-based biomedical image segmentation.
\newblock \emph{Nature methods}, 2021.

\bibitem[Isola et~al.(2017)Isola, Zhu, Zhou, and Efros]{isola2017image}
Phillip Isola, Jun-Yan Zhu, Tinghui Zhou, and Alexei~A Efros.
\newblock Image-to-image translation with conditional adversarial networks.
\newblock In \emph{CVPR}, 2017.

\bibitem[Jakobi et~al.(1995)Jakobi, Husbands, and Harvey]{jakobi1995noise}
Nick Jakobi, Phil Husbands, and Inman Harvey.
\newblock Noise and the reality gap: The use of simulation in evolutionary robotics.
\newblock In \emph{Advances in Artificial Life: Third European Conference on Artificial Life Granada}, 1995.

\bibitem[Jog and Fischl(2018)]{jog2018pulse}
Amod Jog and Bruce Fischl.
\newblock Pulse sequence resilient fast brain segmentation.
\newblock In \emph{MICCAI}, 2018.

\bibitem[Kamnitsas et~al.(2017)Kamnitsas, Baumgartner, Ledig, Newcombe, Simpson, Kane, Menon, Nori, Criminisi, Rueckert, et~al.]{kamnitsas2017unsupervised}
Konstantinos Kamnitsas, Christian Baumgartner, Christian Ledig, Virginia Newcombe, Joanna Simpson, Andrew Kane, David Menon, Aditya Nori, Antonio Criminisi, Daniel Rueckert, et~al.
\newblock Unsupervised domain adaptation in brain lesion segmentation with adversarial networks.
\newblock In \emph{IPMI}, 2017.

\bibitem[Karani et~al.(2021)Karani, Erdil, Chaitanya, and Konukoglu]{karani2021test}
Neerav Karani, Ertunc Erdil, Krishna Chaitanya, and Ender Konukoglu.
\newblock Test-time adaptable neural networks for robust medical image segmentation.
\newblock \emph{MedIA}, 2021.

\bibitem[Kingma and Ba(2014)]{kingma2014adam}
Diederik~P Kingma and Jimmy Ba.
\newblock Adam: A method for stochastic optimization.
\newblock \emph{arXiv preprint arXiv:1412.6980}, 2014.

\bibitem[Kingma and Dhariwal(2018)]{kingma2018glow}
Durk~P Kingma and Prafulla Dhariwal.
\newblock Glow: Generative flow with invertible 1x1 convolutions.
\newblock In \emph{NeurIPS}, 2018.

\bibitem[Krizhevsky et~al.(2012)Krizhevsky, Sutskever, and Hinton]{krizhevsky2012imagenet}
Alex Krizhevsky, Ilya Sutskever, and Geoffrey~E Hinton.
\newblock Imagenet classification with deep convolutional neural networks.
\newblock In \emph{NeurIPS}, 2012.

\bibitem[LaMontagne et~al.(2019)LaMontagne, Benzinger, Morris, Keefe, Hornbeck, Xiong, Grant, Hassenstab, Moulder, Vlassenko, et~al.]{lamontagne2019oasis}
Pamela~J LaMontagne, Tammie~LS Benzinger, John~C Morris, Sarah Keefe, Russ Hornbeck, Chengjie Xiong, Elizabeth Grant, Jason Hassenstab, Krista Moulder, Andrei~G Vlassenko, et~al.
\newblock {OASIS-3}: longitudinal neuroimaging, clinical, and cognitive dataset for normal aging and alzheimer disease.
\newblock \emph{medrxiv}, 2019.

\bibitem[Li et~al.(2021)Li, Yang, Kreis, Torralba, and Fidler]{li2021semantic}
Daiqing Li, Junlin Yang, Karsten Kreis, Antonio Torralba, and Sanja Fidler.
\newblock Semantic segmentation with generative models: Semi-supervised learning and strong out-of-domain generalization.
\newblock In \emph{CVPR}, 2021.

\bibitem[Lim et~al.(2019)Lim, Kim, Kim, Kim, and Kim]{lim2019fast}
Sungbin Lim, Ildoo Kim, Taesup Kim, Chiheon Kim, and Sungwoong Kim.
\newblock Fast autoaugment.
\newblock In \emph{NeurIPS}, 2019.

\bibitem[Liu et~al.(2022)Liu, Deng, Tao, Chu, Duan, and Li]{liu2022undoing}
Yahao Liu, Jinhong Deng, Jiale Tao, Tong Chu, Lixin Duan, and Wen Li.
\newblock Undoing the damage of label shift for cross-domain semantic segmentation.
\newblock In \emph{CVPR}, 2022.

\bibitem[Long et~al.(2015)Long, Shelhamer, and Darrell]{long2015fully}
Jonathan Long, Evan Shelhamer, and Trevor Darrell.
\newblock Fully convolutional networks for semantic segmentation.
\newblock In \emph{CVPR}, 2015.

\bibitem[Maclaurin et~al.(2015)Maclaurin, Duvenaud, and Adams]{maclaurin2015gradient}
Dougal Maclaurin, David Duvenaud, and Ryan Adams.
\newblock Gradient-based hyperparameter optimization through reversible learning.
\newblock In \emph{ICML}, 2015.

\bibitem[Mahmood et~al.(2019)Mahmood, Borders, Chen, McKay, Salimian, Baras, and Durr]{mahmood2019deep}
Faisal Mahmood, Daniel Borders, Richard~J Chen, Gregory~N McKay, Kevan~J Salimian, Alexander Baras, and Nicholas~J Durr.
\newblock Deep adversarial training for multi-organ nuclei segmentation in histopathology images.
\newblock \emph{TMI}, 2019.

\bibitem[Maleki et~al.(2022)Maleki, Ovens, Gupta, Reinhold, Spatz, and Forghani]{maleki2022generalizability}
Farhad Maleki, Katie Ovens, Rajiv Gupta, Caroline Reinhold, Alan Spatz, and Reza Forghani.
\newblock Generalizability of machine learning models: quantitative evaluation of three methodological pitfalls.
\newblock \emph{Radiology: Artificial Intelligence}, 2022.

\bibitem[Milletari et~al.(2016)Milletari, Navab, and Ahmadi]{milletari2016v}
Fausto Milletari, Nassir Navab, and Seyed-Ahmad Ahmadi.
\newblock V-net: Fully convolutional neural networks for volumetric medical image segmentation.
\newblock In \emph{3DV}, 2016.

\bibitem[M{\"u}ller et~al.(2019)M{\"u}ller, Kornblith, and Hinton]{muller2019does}
Rafael M{\"u}ller, Simon Kornblith, and Geoffrey~E Hinton.
\newblock When does label smoothing help?
\newblock In \emph{NeurIPS}, 2019.

\bibitem[Nichol and Dhariwal(2021)]{nichol2021improved}
Alexander~Quinn Nichol and Prafulla Dhariwal.
\newblock Improved denoising diffusion probabilistic models.
\newblock In \emph{ICML}, 2021.

\bibitem[Nikolenko(2021)]{nikolenko2021synthetic}
Sergey~I Nikolenko.
\newblock \emph{Synthetic data for deep learning}.
\newblock Springer, 2021.

\bibitem[Noh et~al.(2015)Noh, Hong, and Han]{noh2015learning}
Hyeonwoo Noh, Seunghoon Hong, and Bohyung Han.
\newblock Learning deconvolution network for semantic segmentation.
\newblock In \emph{ICCV}, 2015.

\bibitem[Papamakarios et~al.(2017)Papamakarios, Pavlakou, and Murray]{papamakarios2017masked}
George Papamakarios, Theo Pavlakou, and Iain Murray.
\newblock Masked autoregressive flow for density estimation.
\newblock In \emph{NeurIPS}, 2017.

\bibitem[Puonti et~al.(2016)Puonti, Iglesias, and Van~Leemput]{puonti2016fast}
Oula Puonti, Juan~Eugenio Iglesias, and Koen Van~Leemput.
\newblock Fast and sequence-adaptive whole-brain segmentation using parametric bayesian modeling.
\newblock \emph{NeuroImage}, 2016.

\bibitem[Richter et~al.(2016)Richter, Vineet, Roth, and Koltun]{richter2016playing}
Stephan~R Richter, Vibhav Vineet, Stefan Roth, and Vladlen Koltun.
\newblock Playing for data: Ground truth from computer games.
\newblock In \emph{ECCV}, 2016.

\bibitem[Ronneberger et~al.(2015)Ronneberger, Fischer, and Brox]{ronneberger2015u}
Olaf Ronneberger, Philipp Fischer, and Thomas Brox.
\newblock {U}-net: Convolutional networks for biomedical image segmentation.
\newblock In \emph{MICCAI}, 2015.

\bibitem[Sandfort et~al.(2019)Sandfort, Yan, Pickhardt, and Summers]{sandfort2019data}
Veit Sandfort, Ke Yan, Perry~J Pickhardt, and Ronald~M Summers.
\newblock Data augmentation using generative adversarial networks (cyclegan) to improve generalizability in {CT} segmentation tasks.
\newblock \emph{Scientific reports}, 2019.

\bibitem[Schulz et~al.(2020)Schulz, Yeo, Vogelstein, Mourao-Miranada, Kather, Kording, Richards, and Bzdok]{schulz2020different}
Marc-Andre Schulz, BT~Thomas Yeo, Joshua~T Vogelstein, Janaina Mourao-Miranada, Jakob~N Kather, Konrad Kording, Blake Richards, and Danilo Bzdok.
\newblock Different scaling of linear models and deep learning in ukbiobank brain images versus machine-learning datasets.
\newblock \emph{Nature communications}, 2020.

\bibitem[Shorten and Khoshgoftaar(2019)]{shorten2019survey}
Connor Shorten and Taghi~M Khoshgoftaar.
\newblock A survey on image data augmentation for deep learning.
\newblock \emph{Journal of big data}, 2019.

\bibitem[Shrivastava et~al.(2017)Shrivastava, Pfister, Tuzel, Susskind, Wang, and Webb]{shrivastava2017learning}
Ashish Shrivastava, Tomas Pfister, Oncel Tuzel, Joshua Susskind, Wenda Wang, and Russell Webb.
\newblock Learning from simulated and unsupervised images through adversarial training.
\newblock In \emph{CVPR}, 2017.

\bibitem[Simonyan and Zisserman(2014)]{simonyan2014very}
Karen Simonyan and Andrew Zisserman.
\newblock Very deep convolutional networks for large-scale image recognition.
\newblock \emph{arXiv preprint arXiv:1409.1556}, 2014.

\bibitem[Sindel et~al.(2023)Sindel, Maier, and Christlein]{sindel2023vesselsegmentation}
Aline Sindel, Andreas Maier, and Vincent Christlein.
\newblock A vessel segmentation-based {CycleGAN} for unpaired multi-modal retinal image synthesis.
\newblock In \emph{BVM Workshop}, 2023.

\bibitem[Sudre et~al.(2017)Sudre, Li, Vercauteren, Ourselin, and Jorge~Cardoso]{sudre2017generalised}
Carole~H Sudre, Wenqi Li, Tom Vercauteren, Sebastien Ourselin, and M Jorge~Cardoso.
\newblock Generalised dice overlap as a deep learning loss function for highly unbalanced segmentations.
\newblock In \emph{Deep Learning in Medical Image Analysis and Multimodal Learning for Clinical Decision Support: Third International Workshop, DLMIA 2017, and 7th International Workshop, ML-CDS 2017, Held in Conjunction with MICCAI 2017}, 2017.

\bibitem[Sushko et~al.(2023)Sushko, Zhang, Gall, and Khoreva]{sushko2023one}
Vadim Sushko, Dan Zhang, Juergen Gall, and Anna Khoreva.
\newblock One-shot synthesis of images and segmentation masks.
\newblock In \emph{WACV}, 2023.

\bibitem[Szegedy et~al.(2015)Szegedy, Liu, Jia, Sermanet, Reed, Anguelov, Erhan, Vanhoucke, and Rabinovich]{szegedy2015going}
Christian Szegedy, Wei Liu, Yangqing Jia, Pierre Sermanet, Scott Reed, Dragomir Anguelov, Dumitru Erhan, Vincent Vanhoucke, and Andrew Rabinovich.
\newblock Going deeper with convolutions.
\newblock In \emph{CVPR}, 2015.

\bibitem[Taylor and Nitschke(2018)]{taylor2018improving}
Luke Taylor and Geoff Nitschke.
\newblock Improving deep learning with generic data augmentation.
\newblock In \emph{SSCI}, 2018.

\bibitem[Tobin et~al.(2017)Tobin, Fong, Ray, Schneider, Zaremba, and Abbeel]{tobin2017domain}
Josh Tobin, Rachel Fong, Alex Ray, Jonas Schneider, Wojciech Zaremba, and Pieter Abbeel.
\newblock Domain randomization for transferring deep neural networks from simulation to the real world.
\newblock In \emph{IROS}, 2017.

\bibitem[Tremblay et~al.(2018)Tremblay, Prakash, Acuna, Brophy, Jampani, Anil, To, Cameracci, Boochoon, and Birchfield]{tremblay2018training}
Jonathan Tremblay, Aayush Prakash, David Acuna, Mark Brophy, Varun Jampani, Cem Anil, Thang To, Eric Cameracci, Shaad Boochoon, and Stan Birchfield.
\newblock Training deep networks with synthetic data: Bridging the reality gap by domain randomization.
\newblock In \emph{CVPR workshops}, 2018.

\bibitem[Tritrong et~al.(2021)Tritrong, Rewatbowornwong, and Suwajanakorn]{tritrong2021repurposing}
Nontawat Tritrong, Pitchaporn Rewatbowornwong, and Supasorn Suwajanakorn.
\newblock Repurposing {GANs} for one-shot semantic part segmentation.
\newblock In \emph{CVPR}, 2021.

\bibitem[Varoquaux(2018)]{varoquaux2018cross}
Ga{\"e}l Varoquaux.
\newblock Cross-validation failure: Small sample sizes lead to large error bars.
\newblock \emph{Neuroimage}, 2018.

\bibitem[Varoquaux and Cheplygina(2022)]{varoquaux2022machine}
Ga{\"e}l Varoquaux and Veronika Cheplygina.
\newblock Machine learning for medical imaging: methodological failures and recommendations for the future.
\newblock \emph{NPJ digital medicine}, 2022.

\bibitem[Wang and Deng(2018)]{wang2018deep}
Mei Wang and Weihong Deng.
\newblock Deep visual domain adaptation: A survey.
\newblock \emph{Neurocomputing}, 2018.

\bibitem[Yarats et~al.(2021)Yarats, Kostrikov, and Fergus]{yarats2021image}
Denis Yarats, Ilya Kostrikov, and Rob Fergus.
\newblock Image augmentation is all you need: Regularizing deep reinforcement learning from pixels.
\newblock In \emph{ICLR}, 2021.

\bibitem[You et~al.(2022{\natexlab{a}})You, Xiang, Su, Zhang, Dong, Onofrey, Staib, and Duncan]{you2022incremental}
Chenyu You, Jinlin Xiang, Kun Su, Xiaoran Zhang, Siyuan Dong, John Onofrey, Lawrence Staib, and James~S Duncan.
\newblock Incremental learning meets transfer learning: Application to multi-site prostate {MRI} segmentation.
\newblock In \emph{International Workshop on Distributed, Collaborative, and Federated Learning}, 2022{\natexlab{a}}.

\bibitem[You et~al.(2022{\natexlab{b}})You, Zhou, Zhao, Staib, and Duncan]{you2022simcvd}
Chenyu You, Yuan Zhou, Ruihan Zhao, Lawrence Staib, and James~S Duncan.
\newblock Simcvd: Simple contrastive voxel-wise representation distillation for semi-supervised medical image segmentation.
\newblock \emph{TMI}, 2022{\natexlab{b}}.

\bibitem[Zhang et~al.(2020)Zhang, Wang, Yang, Sanford, Harmon, Turkbey, Wood, Roth, Myronenko, Xu, et~al.]{zhang2020generalizing}
Ling Zhang, Xiaosong Wang, Dong Yang, Thomas Sanford, Stephanie Harmon, Baris Turkbey, Bradford~J Wood, Holger Roth, Andriy Myronenko, Daguang Xu, et~al.
\newblock Generalizing deep learning for medical image segmentation to unseen domains via deep stacked transformation.
\newblock \emph{TMI}, 2020.

\bibitem[Zhang et~al.(2017)Zhang, Yang, Chen, Fredericksen, Hughes, and Chen]{zhang2017deep}
Yizhe Zhang, Lin Yang, Jianxu Chen, Maridel Fredericksen, David~P Hughes, and Danny~Z Chen.
\newblock Deep adversarial networks for biomedical image segmentation utilizing unannotated images.
\newblock In \emph{MICCAI}, 2017.

\bibitem[Zhang et~al.(2021{\natexlab{a}})Zhang, Ling, Gao, Yin, Lafleche, Barriuso, Torralba, and Fidler]{zhang2021datasetgan}
Yuxuan Zhang, Huan Ling, Jun Gao, Kangxue Yin, Jean-Francois Lafleche, Adela Barriuso, Antonio Torralba, and Sanja Fidler.
\newblock {DatasetGAN}: Efficient labeled data factory with minimal human effort.
\newblock In \emph{CVPR}, 2021{\natexlab{a}}.

\bibitem[Zhang et~al.(2018)Zhang, Yang, and Zheng]{zhang2018translating}
Zizhao Zhang, Lin Yang, and Yefeng Zheng.
\newblock Translating and segmenting multimodal medical volumes with cycle-and shape-consistency generative adversarial network.
\newblock In \emph{CVPR}, 2018.

\bibitem[Zhang et~al.(2021{\natexlab{b}})Zhang, Gao, and Sabuncu]{zhang2021ex}
Zhilu Zhang, Vianne~R Gao, and Mert~R Sabuncu.
\newblock Ex uno plures: Splitting one model into an ensemble of subnetworks.
\newblock \emph{arXiv preprint arXiv:2106.04767}, 2021{\natexlab{b}}.

\bibitem[Zhao et~al.(2019)Zhao, Balakrishnan, Durand, Guttag, and Dalca]{zhao2019data}
Amy Zhao, Guha Balakrishnan, Fredo Durand, John~V Guttag, and Adrian~V Dalca.
\newblock Data augmentation using learned transformations for one-shot medical image segmentation.
\newblock In \emph{CVPR}, 2019.

\bibitem[Zhu et~al.(2017)Zhu, Park, Isola, and Efros]{zhu2017unpaired}
Jun-Yan Zhu, Taesung Park, Phillip Isola, and Alexei~A Efros.
\newblock Unpaired image-to-image translation using cycle-consistent adversarial networks.
\newblock In \emph{ICCV}, 2017.

\end{thebibliography}
}

\clearpage

\twocolumn[
\centering
\Large
\textbf{\emph{Learn2Synth}: Learning Optimal Data Synthesis Using Hypergradients\\for Brain Image Segmentation} \\\vspace{0.05cm}{--- Supplementary Material ---}
\\
\vspace{1.5em}
]

\maketitle

\setcounter{section}{5}
\setcounter{figure}{4}
\setcounter{table}{6}

In the supplementary material, we begin with the related work in \Cref{sec:related}, followed by the details of the datasets in \Cref{sec:datasets} and the experimental details in \Cref{sec:exp_detail}. Next, we provide more results in \Cref{sec:validation}, \Cref{sec:unpair}, \Cref{sec:nnunet} and \Cref{sec:same_test_size}, followed by computational cost in \Cref{sec:cost}. Finally, we discuss the generalization to 3D in \Cref{sec:3d} and the limitations in \Cref{sec:Limitations}.

\section{Related work}
\label{sec:related}
\myparagraph{Deep learning based medical image segmentation.} In the last decades, deep learning methods (CNNs) have provided state-of-the-art accuracy in (medical) image segmentation~\cite{ronneberger2015u, long2015fully, chen2014semantic, chen2018deeplab, chen2017rethinking, noh2015learning}. The UNet architecture~\cite{ronneberger2015u} and its variants~\cite{isensee2021nnu, chen2021transunet} has been one of the most popular methods for medical image segmentation. FCN~\cite{long2015fully} transforms classification CNNs~\cite{krizhevsky2012imagenet, simonyan2014very, he2016deep} to fully-convolutional NNs by replacing fully connected layers with fully convolutional layers. By doing this, FCN transfers the success of classification tasks~\cite{krizhevsky2012imagenet,simonyan2014very,szegedy2015going} to segmentation tasks. Deeplab methods (v1-v2)~\cite{chen2014semantic, chen2018deeplab} add another fully connected Conditional Random Field (CRF) after the last CNN layer to make use of global information instead of using CRF as post-processing. Moreover, dilated/atrous convolutions were introduced in Deeplab v3~\cite{chen2017rethinking} to increase the receptive field and make better use of context information, resulting in better performance.

While deep learning-based methods have achieved impressive performance metrics, they suffer from two major issues. First, deep learning methods usually require a large amount of high-quality labeled data, which is not realistic in scenarios where domain knowledge is needed to obtain the training data. The second issue is the gaps between different domains. The models trained on one domain do not generalize well to other domains that are different from the training data, which is a major problem in medical imaging due to differences in imaging device vendors, imaging protocols, etc.

To solve both issues aforementioned, in this paper, we propose learning a trainable network to augment the synthetic images, which are then used to train a segmentation network, avoiding the requirements of a large amount of training data and overfitting the training data. Due to the power of UNet for image segmentation with fine structures, in this work, we use UNet as a baseline and our backbone network.

\section{Details of the datasets}
\label{sec:datasets}
\myparagraph{ABIDE dataset.} The Autism Brain Imaging Data Exchange (ABIDE)~\cite{di2014autism} is a large, publicly available dataset aimed at advancing the understanding of the intrinsic brain architecture in autism. This dataset contains neuroimaging data from individuals with autism spectrum disorder (ASD) as well as typically developing controls, collected from multiple sites over the world. ABIDE includes structural MRI, resting-state fMRI, and other neuroimaging modalities, alongside extensive demographic and clinical information. The whole dataset contains 1087 younger, high-resolution, isotropic, T1 scans.

\myparagraph{OASIS3 dataset.} The OASIS3~\cite{lamontagne2019oasis} (Open Access Series of Imaging Studies) dataset is a comprehensive, longitudinal collection of neuroimaging, clinical, and cognitive data designed to advance the understanding of normal aging and Alzheimer’s disease (AD). The dataset includes MRI scans, neuropsychological assessments, and clinical evaluations from a diverse cohort of participants, ranging from cognitively healthy individuals to those diagnosed with mild cognitive impairment (MCI) and Alzheimer’s disease. The whole dataset contains 1235 older, high-resolution, isotropic, T1 scans.

\setlength{\tabcolsep}{5pt}
\begin{table*}[ht]
    \centering
    \scriptsize
    \begin{tabular}{l c c c c c c}
        \hline
          Setting & \# of Train, Val., Test & Test Set (MPRAGE) & $3^{\circ}$ & $5^{\circ}$ & $20^{\circ}$ & $30^{\circ}$ \\
        \hline
        SynthSeg & / & $0.861 \pm 0.028$ & $0.776 \pm 0.012$ & $0.694 \pm 0.027$ & $0.766 \pm 0.018$ & $0.781 \pm 0.016$ \\
        \hline
        Supervised UNet (w/ validation) & 29, 5, 5 & $\mathbf{0.941 \pm 0.002}$ & $0.419 \pm 0.025$ & $0.396 \pm 0.020$ & $0.671 \pm 0.026$ & $0.769 \pm 0.022$ \\
        Supervised UNet (w/o validation) & 29, 5, 5 & $0.936 \pm 0.004$ & $0.301 \pm 0.024$ & $0.314 \pm 0.022$ & $0.527 \pm 0.028$ & $0.637 \pm 0.035$ \\
        Supervised UNet (w/ validation) & 5, 5, 29 & $0.907 \pm 0.007$ & $0.397 \pm 0.018$ & $0.413 \pm 0.016$ & $0.586 \pm 0.033$ & $0.692 \pm 0.033$ \\
        Supervised UNet (w/o validation) & 5, 5, 29 & $0.900 \pm 0.011$ & $0.342 \pm 0.019$ & $0.379 \pm 0.020$ & $0.637 \pm 0.032$ & $0.728 \pm 0.023$ \\
        Supervised UNet (w/ validation) & 1, 5, 33 & $0.885 \pm 0.010$ & $0.267 \pm 0.019$ & $0.247 \pm 0.013$ & $0.544 \pm 0.026$ & $0.651 \pm 0.025$ \\
        Supervised UNet (w/o validation) & 1, 5, 33 & $0.871 \pm 0.013$ & $0.337 \pm 0.027$ & $0.372 \pm 0.032$ & $0.654 \pm 0.018$ & $0.727 \pm 0.017$ \\
        \hline
        \emph{Learn2Synth} (w/ validation) & 29, 5, 5 & $0.895 \pm 0.011$ & $\mathbf{0.804 \pm 0.014}$ & $\mathbf{0.789 \pm 0.010}$ & $0.785 \pm 0.025$ & $0.797 \pm 0.023$ \\
        \emph{Learn2Synth} (w/o validation) & 29, 5, 5 & $0.897 \pm 0.012$ & $0.801 \pm 0.012$ & $0.787 \pm 0.011$ & $0.782 \pm 0.024$ & $0.794 \pm 0.023$ \\
        \emph{Learn2Synth} (w/ validation) & 5, 5, 29 & $0.867 \pm 0.030$ & $0.798 \pm 0.013$ & $\mathbf{0.789 \pm 0.010}$ & $\mathbf{0.795 \pm 0.026}$ & $\mathbf{0.799 \pm 0.024}$ \\
        \emph{Learn2Synth} (w/o validation) & 5, 5, 29 & $0.864 \pm 0.031$ & $0.800 \pm 0.014$ & $0.788 \pm 0.012$ & $0.786 \pm 0.024$ & $0.793 \pm 0.023$ \\
        \emph{Learn2Synth} (w/ validation) & 1, 5, 33 & $0.718 \pm 0.045$ & $0.606 \pm 0.020$ & $0.603 \pm 0.017$ & $0.690 \pm 0.036$ & $0.707 \pm 0.032$ \\
        \emph{Learn2Synth} (w/o validation) & 1, 5, 33 & $0.725 \pm 0.037$ & $0.588 \pm 0.024$ & $0.591 \pm 0.018$ & $0.695 \pm 0.035$ & $0.715 \pm 0.031$ \\
        \hline
    \end{tabular}
        \caption{Performance comparison of different models across various datasets.}
    \label{tab:results}
\end{table*}

\myparagraph{Buckner39 dataset.} The Buckner39 dataset~\cite{fischl2002whole} is a comprehensive collection of high-resolution neuroimaging data designed to facilitate the automated labeling and segmentation of neuroanatomical structures within the human brain. The dataset includes T1-weighted MRI scans from 39 healthy adult participants, providing detailed anatomical representations of the brain's major regions. Buckner39 is particularly valuable for evaluating and refining automated segmentation algorithms, offering a benchmark for the accurate identification and labeling of key brain structures across individuals.

\myparagraph{Freesurfer maintenance dataset}~\cite{fischl2004sequence} contains images acquired in 8 subjects with a FLASH sequence at multiple flip angles. In contrast with MPRAGE, the FLASH sequence does not apply an inversion pulse, which results in lower cortical contrast. Furthermore, different flip angles give rise to different contrasts. We use FLASH scans acquired with flip angles of $3^\circ$ and $5^\circ$ (proton density-weighted) and $20^\circ$ and $30^\circ$ (T1-weighted), with the latter being closer in appearance to MPRAGE scans than the former.  All scans were skull-stripped and manually delineated with the same protocol as in~\cite{fischl2002whole}.

\myparagraph{Preprocessing of the datasets.} The original subjects are in 3D space. We first map the original images and the corresponding masks to the 2D atlas space with the command \texttt{mri\_convert}: \url{https://surfer.nmr.mgh.harvard.edu/fswiki/mri_convert}. Then we use \texttt{surfa} (\url{https://surfer.nmr.mgh.harvard.edu/docs/surfa/}) package to map all the labels to 4 unique classes.

\section{Experimental details}
\label{sec:exp_detail}
\myparagraph{Architecture.} We use the standard U-Net~\cite{ronneberger2015u} as our segmentation backbone. It consists of four resolution levels, where each level contains two convolutional layers (each with a $3\times3$ convolution followed by a ReLU activation), followed by a $2\times2$ max pooling operation in the encoder or a transposed convolution (upconvolution) in the decoder.

All experiments use 2D coronal slices extracted from 3D brain MR images. We use the \emph{cornucopia} package\footnote{\url{https://github.com/balbasty/cornucopia}} as our synthetic generator and use a standard UNet~\cite{ronneberger2015u} as the backbone for our segmentation and nonparametric augmentation networks. Architecture details are provided as supplementary material. The networks are randomly initialized and trained from scratch. We use the soft Dice loss~\cite{sudre2017generalised} to supervise the training of the segmentation network. We use the Adam optimizer~\cite{kingma2014adam} with a learning rate of $1\times 10^{-3}$. We apply random intensity augmentations (smoothing, bias field, and noise) and spatial transformations (affine + elastic) to all baselines.

\myparagraph{Dataset details for synthetic experiments in Sec. 4.1.} We used 60\% of the samples as the training set, 20\% as the validation set, and the remaining 20\% as the test set. At each iteration, noise-free synthetic images are generated from these label maps by assigning random intensities to each label. New images are generated at every epoch, yielding a virtually infinite number of synthetic pairs. Overall results are therefore minimally affected by the portion of training samples.

\myparagraph{Dataset details for real-world experiments in Sec. 4.2.1.} We used 60\% of the samples as the training set, 20\% as the validation set, and the remaining 20\% as the test set. Both datasets are multicentric; ABIDE aimed to study participants in the autism spectrum, with an age range slightly biased towards younger participants; OASIS aimed to study patients with dementia, with an age range slightly biased towards older participants. Both datasets mostly contain images acquired with an MPRAGE sequence, which is T1-weighted and optimized for cortical contrast.

\myparagraph{Dataset details for real-world experiments in Sec. 4.2.2.} We use FLASH scans acquired with flip angles of $3^\circ$ and $5^\circ$ (proton density-weighted) and $20^\circ$ and $30^\circ$ (T1-weighted), with the latter being closer in appearance to MPRAGE scans than the former.  All scans were skull-stripped and manually delineated with the same protocol as in~\cite{fischl2002whole}. The synthetic portion of the training set used label maps derived from the ABIDE and OASIS3 datasets.

\section{More results for the generalizability of \emph{Learn2Synth}}
\label{sec:validation}
To further quantify the impact of this validation step, we also provide results obtained with fully converged models as well as the best models selected by using the validation set in~\Cref{tab:results}. 

By comparing the performances of using the validation set or not, we have the following observations:

\begin{enumerate}
    \item There are significant dice point differences on FLASH data between `Best' and `Last' under the same setting, suggesting we will need a validation set for the supervised method.

    \item Our proposed \emph{Learn2Synth} is essentially insensitive to using a validation set or not. So \emph{Learn2Synth} only really needs 5 label examples to achieve satisfactory performance, instead of 10 (5 training samples plus 5 validation samples).
\end{enumerate}

\section{Unpaired segmentation}
\label{sec:unpair}
An advantage of SynthSeg~\cite{billot2023synthseg} is that it performs unpaired segmentation, which only requires a set of segmentation maps for MRI synthesis during segmentation training. In contrast, our proposed \emph{Learn2Synth} framework requires a small amount of labeled real data during training to learn the synthesis process.

\begin{figure}[h]
\centering
    \includegraphics[width=0.47\textwidth]{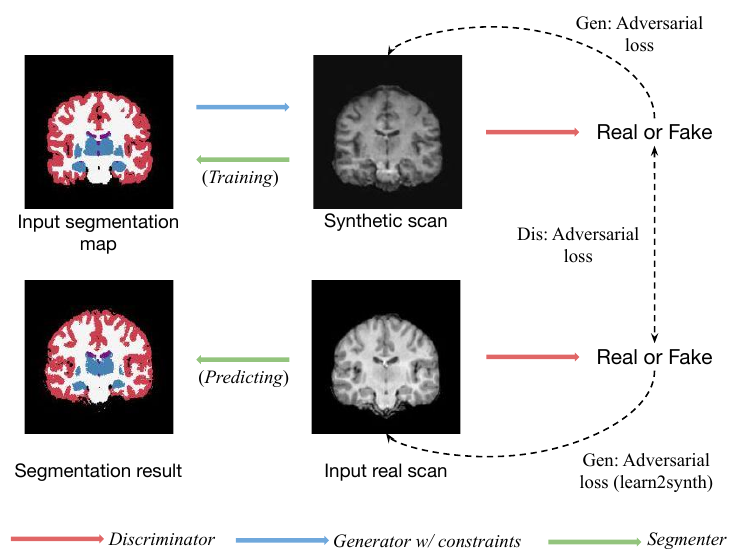}
    \caption{Illustration of the unpaired approach for learning segmentation with \emph{Learn2Synth}. The generator is trained with a unique adversarial loss that simultaneously aims to make synthetic scans more realistic while making real scans appear less authentic to the discriminator through \emph{Learn2Synth} hypergradient back-propagation. This approach does not require paired segmentation-label data, as the input segmentation maps and real scans are independent of each other.}
    \label{fig:unpaired}
\end{figure}

Here, we explore the potential of unpaired segmentation using \emph{Learn2Synth} in combination with a Generative Adversarial Network (GAN). As illustrated in Figure \ref{fig:unpaired}, we use a GAN to generate a synthetic MRI scan that is constrained to an input segmentation map. Unlike conventional GANs, the adversarial loss on the generator is applied not only to the synthetic scan but also to the real scan. Specifically, the generator is trained to fool the discriminator by making the fake data appear more realistic, while simultaneously making the real data appear faker, through the \emph{Learn2Synth} hypergradient back-propagation. The input segmentation maps and real scans are independent of each other, meaning that paired segmentation-label data is not required.

We conducted experiments under the same conditions as described in Section 4.2.1. Unpaired segmentation maps from ABIDE and OASIS-3, along with scans from ABIDE, OASIS-3, and Buckner39, were used as training samples. The trained segmenter was then applied to the Buckner39 dataset, resulting in a Dice score of 0.874 ($\pm$ 0.0104), outperforming SynthSeg (0.861), as shown in Table 6.

\section{nnUNet as backbone}
\label{sec:nnunet}
\Cref{table:nnunet} shows results using nnUNet as the seg. backbone for both `Naive SynthSeg' and \textit{Learn2Synth}, with improved performance but much longer training time ($\approx$62.1h vs $\approx$17.7h on OASIS3). Our focus is on training strategies rather than architectures; networks can be treated as black boxes, and contrast invariance cannot be learned if trained only on T1w images.

\begin{table}[ht]
\centering
\scriptsize
\begin{tabular}{ccc}
\hline
Method & ABIDE & OASIS3 \\ \hline
Naive SynthSeg & 0.881 & 0.863 \\
\emph{Learn2Synth} (parametric setting fixed $\sigma$) & \textbf{0.893} & \textbf{0.875}\\
\hline
\end{tabular}
\caption{Comparison using nnUNet as seg. backbone.}
\label{table:nnunet}
\end{table}


\section{Comparison with the same test size and add Mixed SynthSeg as baseline}
\label{sec:same_test_size}
We also reported the results in Table 6 (main text) using a consistent test set size across all comparisons for interpretable evaluation. Additionally, we have included both the `Mixed SynthSeg' and `Finetuned SynthSeg' baselines for comparison (\Cref{tab:supp_manual_results}). We only include the results here for the flip angle of FLASH $3^\circ$ for space limitations, and the others will be included in the revised version.

\begin{table}[!ht]
\setlength{\tabcolsep}{3pt}
    \centering
    \scriptsize
    \begin{tabular}{l c c c}
        \hline
        Setting & \# of Train, Val., Test & Test Set (MPRAGE) & $3^{\circ}$ \\
        \hline
        SynthSeg & / & 0.861 $\pm$ 0.028 & 0.776 $\pm$ 0.012\\
        Mixed SynthSeg & 5, 5, 5 & 0.859 $\pm$ 0.021 & 0.781 $\pm$ 0.013 \\
        Finetuned SynthSeg & 5, 5, 5 & 0.863 $\pm$ 0.017 & 0.783 $\pm$ 0.015 \\
        \hline
        Supervised UNet & 29, 5, 5 & \textbf{0.941 $\pm$ 0.002} & 0.419 $\pm$ 0.025\\
        Supervised UNet & 5, 5, 5 & 0.910 $\pm$ 0.012 & 0.397 $\pm$ 0.018\\
        Supervised UNet & 1, 5, 5 & 0.879 $\pm$ 0.014 & 0.267 $\pm$ 0.019\\
        \hline
        \emph{Learn2Synth} & 29, 5, 5 & 0.895 $\pm$ 0.011 & \textbf{0.804 $\pm$ 0.014} \\
        \emph{Learn2Synth} & 5, 5, 5 & 0.871 $\pm$ 0.028 & 0.798 $\pm$ 0.013 \\
        \emph{Learn2Synth} & 1, 5, 5 & 0.725 $\pm$ 0.019& 0.606 $\pm$ 0.020 \\
        \hline
    \end{tabular}%
        \caption{Comparison of models across various datasets.}
    \label{tab:supp_manual_results}
\end{table}

\section{Computational cost and framework complexity}
\label{sec:cost}
For OASIS3, taking the `\emph{Learn2Synth} (parametric setting with fixed $\sigma$)' as an example, the model converges after 1{,}500 epochs with a batch size of 64, requiring $\approx$17.7 hours of training time on an NVIDIA L40S GPU (48GB), using a 64-core Intel(R) Xeon(R) Gold 6438Y+ CPU and 200 GB RAM. For comparison, `Naive SynthSeg' requires $\approx$10.3 hours under the same setup.

While \textit{Learn2Synth} introduces alternating synthetic and real passes, it eliminates the need for costly cross-validation typically used to tune augmentation hyperparameters. Unlike grid search, which scales exponentially with the number of parameters, \textit{Learn2Synth} uses hypergradient-based updates to optimize parameters in a single training run, making the process far more efficient and scalable.

A key advantage of \textit{Learn2Synth} is that it avoids manual hyperparameter tuning by treating augmentation parameters as learnable variables. Unlike traditional methods like SynthSeg that rely on heuristic tuning, \textit{Learn2Synth} uses hypergradients and a small set of real validation data to automatically optimize augmentations, improving generalizability across domains.

\section{Generalization to 3D data}
\label{sec:3d}
While our experiments use 2D slices, the \textit{Learn2Synth} framework is architecture- and dimensionality-agnostic and readily extends to 3D. Its synthetic-to-real training and hypergradient-based optimization remain applicable in volumetric settings, which pose challenges like anisotropy and high memory demands. We plan to explore full 3D evaluations in future work. 

\section{Limitations}
\label{sec:Limitations}
One limitation compared with naive SynthSeg~\cite{billot2023synthseg} is that we need labeled real scans in the target modality to optimize the segmentation results. Also, this work focuses specifically on segmentation, as indicated by the paper’s scope. While the core ideas of \textit{Learn2Synth} could extend to tasks like registration or lesion detection, we intentionally limit our study to segmentation for a focused evaluation. Exploring other tasks is left for future work.

\end{document}